\documentclass[journal]{IEEEtran}
\usepackage{amsmath,amsfonts}
\usepackage{algorithmic}
\usepackage{algorithm}
\usepackage{array}
\usepackage[caption=false,font=normalsize,labelfont=sf,textfont=sf]{subfig}
\usepackage{textcomp}
\usepackage{bm}
\usepackage{mathrsfs}
\usepackage{amssymb}
\usepackage{stfloats}
\usepackage{url}
\usepackage{verbatim}
\usepackage{graphicx}
\usepackage{xcolor}
\usepackage{booktabs}
\usepackage{multirow}
\usepackage{makecell}
\usepackage{tabularx}
\usepackage{threeparttable}
\usepackage{cite}
\usepackage{hyperref}

\begin{document}

\title{IRNN: Innovation-driven Recurrent Neural Network \\for Time-Series Data Modeling and Prediction}

\author{Yifan~Zhou,
        Yibo~Wang,
        Chao~Shang,~\IEEEmembership{Member,~IEEE}
\thanks{Yifan Zhou and Yibo Wang are with The Department of Automation, Tsinghua University, Beijing 100084, China (e-mail: \href{mailto:zhouyifa24@mails.tsinghua.edu.cn}{zhouyifa24@mails.tsinghua.edu.cn}; \href{mailto:wyb21@mails.tsinghua.edu.cn}{wyb21@mails.tsinghua.edu.cn}).}%
\thanks{Chao Shang is with the Department of Automation, Beijing National Research Center for Information Science and Technology, Tsinghua University, Beijing 100084, China (e-mail: \href{mailto:c-shang@tsinghua.edu.cn}{c-shang@tsinghua.edu.cn}).}}

\markboth{IEEE Transactions on Knowledge and Data Engineering}%
{Shell \MakeLowercase{\textit{et al.}}: A Sample Article Using IEEEtran.cls for IEEE Journals}

\IEEEpubid{0000--0000/00\$00.00~\copyright~2025 IEEE}

\maketitle

\begin{abstract}
Many real-world datasets are time series that are sequentially collected and contain rich temporal information. Thus, a common interest in practice is to capture dynamics of time series and predict their future evolutions. To this end, the recurrent neural network (RNN) has been a prevalent and effective machine learning option, which admits a nonlinear state-space model representation. Motivated by the resemblance between RNN and Kalman filter (KF) for linear state-space models, we propose in this paper Innovation-driven RNN (IRNN), a novel RNN architecture tailored to time-series data modeling and prediction tasks. By adapting the concept of ``innovation'' from KF to RNN, past prediction errors are adopted as additional input signals to update hidden states of RNN and boost prediction performance. Since innovation data depend on network parameters, existing training algorithms for RNN do not apply to IRNN straightforwardly. Thus, a tailored training algorithm dubbed input updating-based back-propagation through time (IU-BPTT) is further proposed, which alternates between updating innovations and optimizing network parameters via gradient descent. Experiments on real-world benchmark datasets show that the integration of innovations into various forms of RNN leads to remarkably improved prediction accuracy of IRNN without increasing the training cost substantially. 
\end{abstract}


\section{Introduction}
\IEEEPARstart{T}{ime} series modeling and prediction have attracted immense attentions from a broad spectrum of fields, including power systems \cite{8039509}, weather prediction \cite{bi2023accurate}, and traffic planning \cite{lim2021time}. In real-world production processes, high-precision prediction of time series enables more strategic decisions to be made subsequently. For instance, in the power sector, short-term electricity demand prediction facilitates the real-time optimization of power generation plans. In dynamic and safety-critical environments such as autonomous driving, reliable short-term prediction allows vehicles to anticipate changes in the surrounding environment and make decisions proactively \cite{fu2024summary}.

The importance of time-series prediction lies in high precision, adaptability, and robustness, and prediction models are pivotal in predicting precise values, trends, or potential occurrences, leveraging historical time series data \cite{wang2024mpm}. For short-term prediction tasks, minimizing prediction error and delay at each step is critical to ensuring timely and reliable prediction \cite{chen2020importance}. In real-world systems, data often contain noise or uncertainties, requiring models to provide relatively accurate and stable prediction even when dealing with fluctuations or anomalies in the input \cite{sangiorgio2021forecasting}. Traditional statistical models, such as auto-regressive integrated moving average (ARIMA), have been heavily used to capture linear dynamics in noisy data and predict future evolutions \cite{nelson1998time}. However, many real-world time-series exhibit pronounced nonlinear characteristics with complex dynamics, rendering conventional methods inadequate for capturing the underlying behaviors of such systems \cite{tealab2018time, kaur2023autoregressive}.

In recent years, the advance of deep neural networks has offered new possibilities for time-series data modeling and prediction. Among various deep structures, the recurrent neural networks (RNNs) play a leading role in the field of time series prediction. Their recurrent structure, which transmits information across sequences, make them inherently well-suited for modeling time-series data of varying lengths. Over decades of development, RNNs have evolved into numerous variants including Long Short-Term Memory (LSTM) and Gated Recurrent Unit (GRU) \cite{hochreiter1997long, cho2014properties}. By employing gating mechanisms to address vanishing gradients, these advanced recurrent architectures have garnered widespread applications across various fields \cite{jia2024witran, shih2019temporal, qin2017dual, wang2022ngcu, qin2024hierarchically}. 

\IEEEpubidadjcol
The recursive structure of RNNs are highly adaptable for integration with other techniques. For example, RNNs can be combined with autoregressive methods to model the probabilistic distribution of time series, thereby providing a framework to predict future fluctuations and evaluate risks in high-uncertainty scenarios \cite{salinas2020deepar}. A hybrid architecture combining RNN with Bayesian inference enables collaborative time-series and event sequence prediction, while maintaining network interpretability \cite{li2022learning}. Enhancements to gating mechanisms, whether through modifying existing gates in classical structures (e.g., LSTM and GRU) or introducing new gates, enable the stack of deep recurrent network \cite{wang2022ngcu, qin2024hierarchically, li2018independently}. In summary, these approaches help to improve the performance of RNNs, highlighting the inherent flexibility and adaptability of the RNN framework. 

Many real-world data manifest as time series containing rich temporal information, while their intrinsic nature stems from observations of complex dynamic systems. To describe linear dynamics, the Kalman Filter (KF) yields an optimal state estimator for linear time-invariant (LTI) systems subject to Gaussian noise \cite{deng2005new}. KF critically relies on prediction errors, also referred to as innovations, to continually refine the state estimation and output prediction  \cite{draper1998applied, oh2012predictive}. In \cite{wang2025data}, a direct data-driven implementation of KF for output prediction was presented, which bears close resemblance with KF for stochastic LTI systems.

Although KF and their variants exhibit excellent performance in output prediction of LTI systems with uncertainties, real-world time series often embody nonlinear dynamics, posing challenges to these linear modeling methods \cite{shao2024exploring}. RNNs, which can be viewed as a class of nonlinear dynamic systems, offer a powerful alternative. While RNNs and KF share similar recurrent structures, existing RNN architectures make no use of prediction errors to correct potential errors in hidden states when handling time-series data. This inspires us to incorporate the concept of innovation from KF into RNNs, such that the recurrence of hidden states becomes aware of past prediction errors and is able to compensate for such errors by itself. In summary, the primary contributions of our work are as follows:

\begin{enumerate}
    \item We propose \textbf{I}nnovation-driven \textbf{RNN} (\textbf{IRNN}), a novel RNN structure specifically designed for time series modeling and prediction. By adapting the concept of innovation from KF, past prediction errors are adopted as additional input signals to update hidden states of RNN and improve prediction accuracy effectively.

    \item The idea of incorporating innovations can be applied to various forms of RNN. In particular, by adding innovations into GRU and LSTM with gating mechanisms, we derive Innovation-driven GRU (IGRU) and Innovation-driven LSTM (ILSTM) that are specifically designed for time-series data modeling and prediction.
    
    \item Traditional back-propagation through time (BPTT) algorithm cannot be directly applied to train IRNNs. Thus we propose Input-Updating BPTT (IU-BPTT), a tailored training algorithm, which alternates between updating innovations and optimizing network parameters via gradient descent. This enables to effectively train IRNNs by modularly invoking the off-the-shelf BPTT routine.
    
    \item We conduct experiments on real-world datasets using the single-layer IRNN, IGRU and ILSTM. Experiment results demonstrate that IRNNs can be trained with IU-BPTT effectively, and significantly outperform their generic counterparts without innovations in multi-step prediction tasks. In addition, these improvements are achieved without noticeably increasing the number of parameters or the training cost.
\end{enumerate}

The remainder of the paper is organized as follows. In Section \ref{sec:RelatedWork} related works are first reviewed. Section \ref{sec:InnoArch} motivates our IRNN and presents a tailored training algorithm. Section \ref{sec:expriment} presents experimental results of IRNNs and IU-BPTT performance, along with ablation studies. Section \ref{sec:conclusion} concludes our work and discusses future research directions.

\section{Connections between RNN and Control Theory}
\label{sec:RelatedWork}
There has been a close connection between the biologically-inspired RNNs and the control systems community. As an effective data-driven modeling tool, RNNs can represent temporal dynamic behavior through feedback loops in neurons and the memory capacity of hidden states. Thus RNNs offer a viable option for modeling nonlinear dynamic systems in control applications \cite{wang2008design, 6029334}. For instance, RNNs have been incorporated in the design of model predictive control (MPC) \cite{wu2019machine, wu2019machine2}. Following this work, a physics-based RNN modeling approach was introduced by combining prior knowledge from the target system with RNN structure to predict future state evolution in the optimization problem of MPC \cite{wu2020process}. Ref. \cite{wang2017new} proposed to use LSTM for modeling and controlling dynamic systems from data. To describe complex dynamics of nonlinear systems, an efficient model was proposed by merging the feed-forward neural network and RNN \cite{shobana2023recurrent}.

On the other hand, control-theoretic results have also inspired new development of deep sequence models. As a key concept in modern control theory, state space models (SSM) exhibit a linear recurrent structure and motivate many new architectures of RNN, e.g., deep SSMs \cite{gu2022efficiently} and Mamba \cite{gu2023mamba}. Specifically, the Mamba introduces a selective mechanism into SSM, which allows network parameters to be nonlinearly dependent on data \cite{wang2025mamba}. Meanwhile, filtering algorithms in the control theory, notably the KF and its variants, also motivate new effective algorithms for training RNNs that can be viewed as dynamical systems. For example, to stabilize the training process of RNNs, an extended KF (EKF)-based training strategy was proposed in \cite{wang2011convergence}, and its effectiveness under general convex loss functions and regularization terms was further investigated in \cite{bemporad2022recurrent}. 

This paper also draws inspiration from control theory to boost the performance of RNNs but differs substantially from these existing works. To be more specific, we focus on the task of time-series data modeling and prediction, and present a new KF-like architecture of RNN.

%The interplay between RNNs and control theory inspires us to integrate methods from KF for addressing uncertainties into RNN architecture design.

\section{Innovation-Driven RNN for Time-Series Data}
\label{sec:InnoArch}
\subsection{Motivation with Innovations}
Consider a time series data trajectory $\mathcal{T} = \{ (\bm{u}_t,\bm{y}_t)\}_{t=1}^{T}$, where $\bm{u}_t \in \mathbb{R}^{n_u}$ and $\bm{y}_t \in \mathbb{R}^{n_y}$ are multivariate exogenous inputs and outputs of a dynamical system, with $n_u$ and $n_y$ denoting the feature dimension of $\bm{u}_t$ and $\bm{y}_t$. Through the recurrence in hidden states, RNNs offer a viable option for modeling the dynamics of $\mathcal{T}$ as well as the causal relation between $\bm{u}_t$ and $\bm{y}_t$. An RNN can be formally expressed as a nonlinear state-space model \cite{rangapuram2018deep}:
\begin{equation}
    \label{eq:generalRNN}
    \left\{
    \begin{aligned}
        \bm{x}_{t} &= f\left( \bm{x}_{t-1}, \bm{u}_{t}, \bm{y}_{t-1}\right),\\
        \hat{\bm{y}}_{t}   &= h\left( \bm{x}_{t}, \bm{u}_{t}\right),
    \end{aligned}
    \right. 
\end{equation}
where $\bm{x}_{t} \in \mathbb{R}^{n_x}$ represents  dynamically evolving hidden states, $\hat{\bm{y}}_t \in \mathbb{R}^{n_y}$ is the predicted output, and $f(\cdot)$ and $h(\cdot)$ are nonlinear functions characterizing forward evolutions of the hidden states and the state-output relation, respectively. The Markovian property of $\bm{x}_{t}$ and the recursive form of state-space model \eqref{eq:generalRNN} enable RNN to succinctly delineate nonlinear dynamics of time-series data and enable faithful prediction of future behavior. As an example of \eqref{eq:generalRNN}, the generic single-layer RNN employs a fully-connected single-layer network to describe the state evolution:
\begin{equation}\label{eq:RNN_1}
\left \{
  \begin{aligned}
    {{\bm{x}}_{t}} &=\sigma({{\mathbf{W}}_{\rm xx}}{{\bm{x}}_{t-1}}+{{\mathbf{W}}_{\rm xu}}{{\bm{u}}_{t}}+{{\mathbf{W}}_{\rm xy}}{{\bm{y}}_{t-1}}+{{\bm{b}}_{\rm x}}),\\
    {{\hat{\bm{y}}}_{t}} &={{\mathbf{W}}_{\rm yx}}{{\bm{x}}_{t}}+{{\bm{b}}_{\rm y}},
  \end{aligned}
\right .
\end{equation}
where $\mathbf{W}_{\rm xx} \in \mathbb{R}^{n_x \times n_x}$, $\mathbf{W}_{\rm xu} \in \mathbb{R}^{n_x \times n_u}$, $\mathbf{W}_{\rm xy} \in \mathbb{R}^{n_y \times n_x}$ are weight matrices of the hidden layer, and $\bm{b}_{\rm x} \in \mathbb{R}^{n_x}$ is the bias vector. $\mathbf{W}_{\rm yx} \in \mathbb{R}^{n_y \times n_x}$ and $\bm{b}_{\rm y} \in \mathbb{R}^{n_y}$ are the weight matrix and bias of the output layer. Common choices for the nonlinear activation function $\sigma(\cdot)$ include the sigmoid function and the hyperbolic tangent function:
\begin{equation}
    \operatorname{Sigmoid}(x) = \frac{1}{1 + e^{-x}}, \quad
    \operatorname{Tanh}(x) = \frac{e^{x} - e^{-x}}{e^{x} + e^{-x}}.
\end{equation}

When tackling LTI systems with Gaussian noise, KF is known to yield the optimal estimator of states and outputs. Consider a standard discrete-time stochastic LTI system:
\begin{equation}
    \label{eq:stochastic_LTI}
    \left\{
    \begin{aligned}
    \bm{x}_{t+1} &= \mathbf{A} \bm{x}_{t} + \mathbf{B} \bm{u}_{t} + \bm{w}_{t}, \\
    \bm{y}_{t} &= \mathbf{C} \bm{x}_{t} + \mathbf{D} \bm{u}_{t} + \bm{v}_{t},
    \end{aligned}
    \right.
\end{equation}
where $\bm{w}_t \in \mathbb{R}^{n_x}$ and $\bm{v}_t \in \mathbb{R}^{n_y}$ are zero-mean white Gaussian process disturbance and measurement noise, respectively, and $\{ \mathbf{A}, \mathbf{B}, \mathbf{C}, \mathbf{D} \}$ are system matrices. In this case, the KF is known to be optimal in terms of estimation error minimization \cite{kamen2012introduction}:
\begin{equation}
\label{eq:KF}
\left\{
\begin{aligned}
\hat{\bm{x}}_{t+1} &= \mathbf{A} \hat{\bm{x}}_{t} + \mathbf{B} \bm{u}_{t} + \mathbf{K} (\bm{y}_{t}-\hat{\bm{y}}_{t}), \\
\hat{\bm{y}}_{t} &= \mathbf{C} \hat{\bm{x}}_{t} + \mathbf{D} \bm{u}_{t},
\end{aligned}
\right.
\end{equation}
where $\hat{\bm{x}}_{t}$ denotes the state estimate and $\mathbf{K}$ is the steady-state Kalman gain matrix. Central to the KF recursion \eqref{eq:KF} is the one-step ahead prediction error $\bm{e}_{t} \triangleq \bm{y}_{t}-\hat{\bm{y}}_{t}$, also called the \textit{innovation}. It essentially implies unmeasurable errors in state estimation $\hat{\bm{x}}_{t}$ and thus can be used to compensate for potential imperfection in $\hat{\bm{x}}_{t}$ via feedback. Incorporating $\bm{e}_{t}$ is a prerequisite to ensure optimality of KF, which is also the case for data-driven implementation of KF \cite{wang2025data}.

Unlike the state recursion in KF \eqref{eq:KF}, the recursive updating rule of RNN in \eqref{eq:generalRNN} is \textit{not} informed of past prediction errors, which can limit its promise in time-series modeling and prediction. Thus, a simple and intuitive idea is to inject meaningful innovation information into RNN architectures so as to better hedge against uncertainty and enhance their capability in time-series data modeling. To our knowledge, this intuition has yet to be seen in current RNN literature.

\subsection{Innovation-Driven Architecture of RNN}
\label{architecture}
\begin{figure}[hbtp]
    \centering
    \subfloat[Single-layer RNN]{
        \includegraphics[width=\columnwidth]{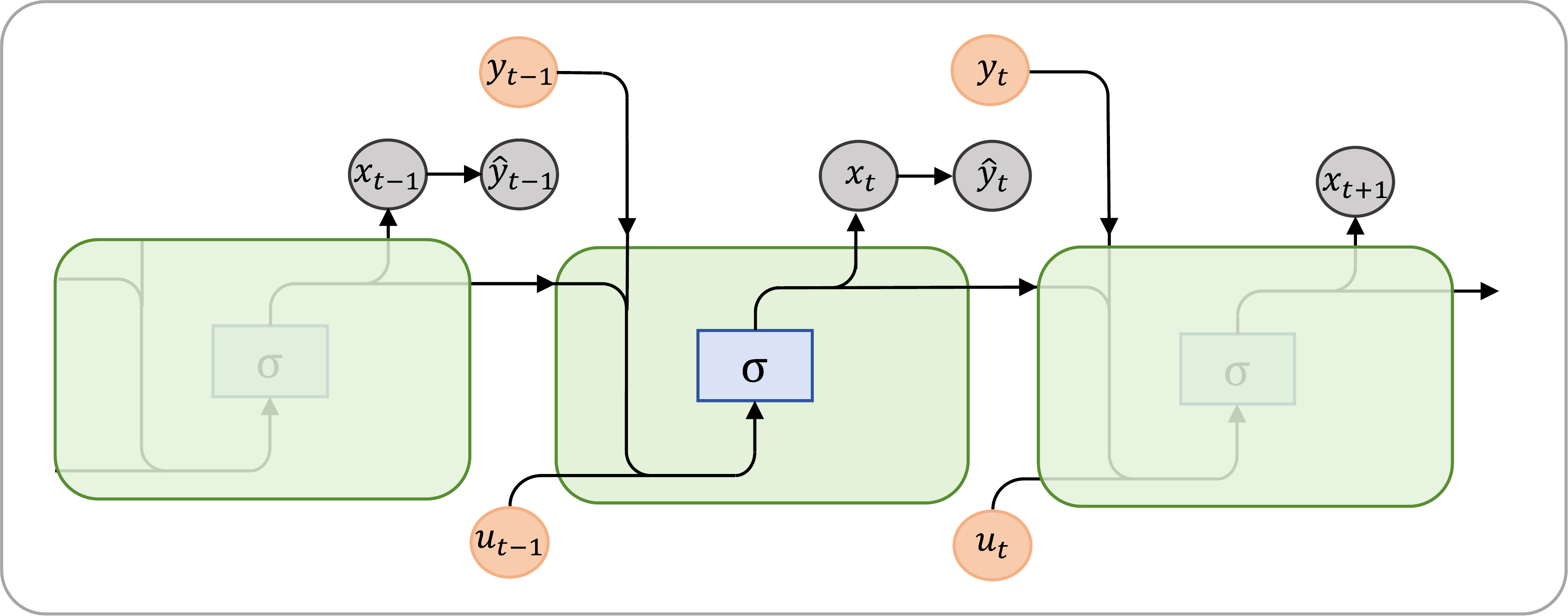}
        \label{fig:RNN_structure}}\\
    \subfloat[Single-layer IRNN]{
        \includegraphics[width=\columnwidth]{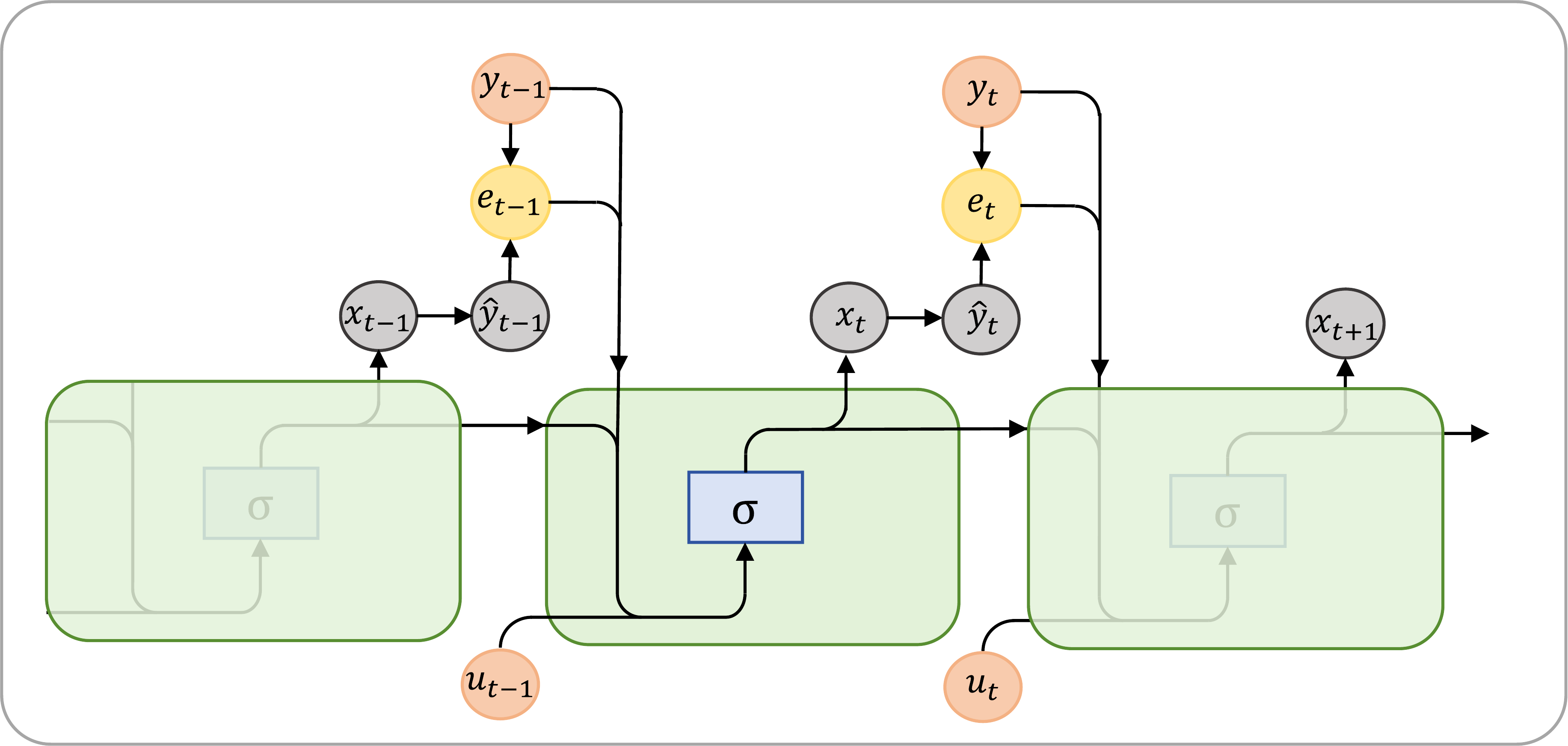}
        \label{fig:IRNN_structure}}
    \caption{Illustration of single-layer RNN (\ref{fig:RNN_structure}) and single-layer IRNN (\ref{fig:IRNN_structure}) architecture for time-series data. The prediction $\hat{\bm{y}}_t$ is made at each time instance with continuous input $\bm{u}_t$ to generate $\bm{e}_t$. The calculation details of \(\bm{x}_t\) are displayed within the yellow rectangle.}
    \label{fig:RNN_structures}
\end{figure}

For time-series data modeling, we propose to incorporate the past innovation $\bm{e}_{t-1} = \bm{y}_{t-1} - \hat{\bm{y}}_{t-1}$ as additional inputs of RNN at current time $t$. This inspires IRNN, a new RNN architecture for effectively handling temporal data. Its general form can be described as:
\begin{equation}
    \label{eq:generalIRNN}
    \left\{
    \begin{aligned}
        \bm{x}_{t} &= f\left( \bm{x}_{t-1}, \bm{u}_{t}, \bm{y}_{t-1}, \bm{e}_{t-1}\right) \\
        \hat{\bm{y}}_{t}   &= h\left( \bm{x}_{t}, \bm{u}_{t}\right) \\
        \bm{e}_{t} & = \bm{y}_t - \hat{\bm{y}}_{t}
    \end{aligned} 
    \right.
\end{equation}
as an extension of \eqref{eq:generalRNN}. 
In a similar spirit, we can embed past innovations into the generic single-layer RNN structure \eqref{eq:KF}, and arrive at a ``Kalman-like" generalization of single-layer IRNN:
\begin{equation}\label{eq:IRNN_1}
\left \{
  \begin{aligned}
    {{\bm{x}}_{t}} =&\sigma({{\mathbf{W}}_{\rm xx}}{{\bm{x}}_{t-1}}+{{\mathbf{W}}_{\rm xu}}{{\bm{u}}_{t}}+{{\mathbf{W}}_{\rm xy}}{{\bm{y}}_{t-1}} +{{\mathbf{W}}_{\rm xe}}{{\bm{e}}_{t-1}}+{{\bm{b}}_{\rm x}}),\\
    \hat{{\bm{y}}}_{t} =&{{\mathbf{W}}_{\rm yx}}{{\bm{x}}_{t}}+{{\bm{b}}_{\rm y}},\\
    {{\bm{e}}_{t}} =&{{\bm{y}}_{t}}-\hat{\bm{y}}_{t},
  \end{aligned} 
  \right .
\end{equation}
where $\mathbf{W}_{\rm xe} \in \mathbb{R}^{n_x \times n_y}$ is a learnable weight matrix that plays a similar role as the Kalman gain $\mathbf{K}$ in \eqref{eq:KF}. A structural comparison between the generic single-layer RNN \eqref{eq:RNN_1} and the corresponding single-layer IRNN \eqref{eq:IRNN_1} is depicted in Fig. \ref{fig:RNN_structures}, where the latter bears a rather different philosophy from the former. At time instance $t$, the true output $\bm{y}_{t-1}$ at previous time instance becomes available and thus the innovation $\bm{e}_{t-1}$ can be immediately computed. Intuitively, a large prediction error $\bm{e}_{t-1}$ is mainly attributed to the unseen error in the hidden state $\bm{x}_{t-1}$, so ${{\bm{e}}_{t-1}}$ can be viewed as an important information carrier of errors in $\bm{x}_{t-1}$. Like KF, the hidden unit of IRNN becomes aware of past errors with ${{\bm{e}}_{t-1}}$ used as inputs and is capable of compensating for such errors through innovation-driven feedback. By contrast, the generic RNN does not carry such a Kalman-like interpretation of error correction. In this sense, the proposed IRNN better parallels KF than the generic RNN and thus may yield improved performance of time-series data modeling. 

Next we proceed to the problem of multi-step ahead prediction. That is, given network parameters, the current hidden state $\bm{x}_{t-1}$ and future inputs $\{\bm{u}_t, \bm{u}_{t+1}, \cdots \}$, how to reasonably predict the resulting outputs $\{\hat{\bm{y}}_t, \hat{\bm{y}}_{t+1}, \cdots\}$. Recall that for the generic single-layer RNN \eqref{eq:RNN_1}, this can be readily achieved by iterating the recurrently updating equation in \eqref{eq:RNN_1} and replacing $\bm{y}_t$ with $\hat{\bm{y}}_t$. This gives rise to:
\begin{equation}\label{eq:RNN_2}
    \begin{split}
        {{\hat{\bm{x}}}_{t+k+1 \mid t}} &=\sigma({{\mathbf{W}}_{\rm xx}}{\hat{\bm x}_{t+k \mid t}}+{{\mathbf{W}}_{\rm xu}}{{\bm{u}}_{t+k+1}}+{{\mathbf{W}}_{\rm xy}}{\hat{{\bm y}}_{t+k \mid t}}+{{\bm{b}}_{\rm x}}),\\
        {\hat{\bm y}_{t+k+1\mid t}} &= \mathbf{W}_{\rm yx} {\hat{\bm x}_{t+k+1 \mid t}}+{{\bm{b}}_{\rm y}},
    \end{split}
\end{equation}
where ${\hat{{\bm x}}_{t+k \mid t}}$ and $\hat{\bm y}_{t+k \mid t}$ are the $k$-step ahead predictions of hidden state and output at current instance $t$. As for IRNN, we can adopt a similar idea but have to account for future innovations that are essentially unknown. In fact, it suffices to set future innovations to zero, inherently assuming that there is no output prediction error in future propagation. This yields the following multi-step ahead predictor based on IRNN:
\begin{equation}\label{eq:IRNN_2}
\begin{split}
    \hat{\bm{x}}_{t+k+1 \mid t} &= \sigma({{\mathbf{W}}_{\rm xx}}{{\hat{{\bm{x}}}_{t+k \mid t}}}+{{\mathbf{W}}_{\rm xu}}{{\bm{u}}_{t+k+1}}\\
    &\quad \quad ~~ +{{\mathbf{W}}_{\rm xy}}{{\hat{\bm{y}}_{t+k\mid t}}}+{{\mathbf{W}}_{\rm xe}}\cdot \mathbf{0}+{{\bm{b}}_{\rm x}}),\\
    \hat{\bm{y}}_{t+k+1\mid t} &= {\mathbf{W}_{\rm yx}}{\hat{\bm{x}}}_{t+k+1 \mid t}+{{\bm{b}}_{\rm y}}.
\end{split}
\end{equation}

Note that IRNNs are specifically designed for time-series prediction tasks and may not be applicable to other sequence tasks. This is because time series prediction inherently allows access to historical observations of the target variable(s) and error correction \textit{at each time instance}. In contrast, for some sequence-to-sequence tasks (e.g., machine translation and text generation), networks recursively depend on self-generated outputs and reference outputs are only available post hoc (if at all), which no longer allows to compute innovations continually.

\subsection{Extension to LSTM and GRU}\label{sec:IGRU and ILSTM}
RNN is known to suffer from gradient vanishing and exploding issues during backpropagation through extended sequence lengths \cite{jozefowicz2015empirical}. To combat these limitations, LSTM was developed as a specialized RNN variant with enhanced capabilities for modeling long-term dependencies \cite{fang2021survey}. The technical novelty of LSTM lies in the data-dependent gating mechanisms for regulating information flow over time. Specifically, there are three gating units in LSTM: the input gate $\bm{g}^{\rm i}_{t}$, forget gate $\bm{g}^{\rm f}_{t}$, and output gate $\bm{g}^{\rm o}_{t}$ defined as:
\begin{equation}\label{eq:LSTM_gates}
    \begin{split}
    {\bm{g}^{\rm f}_{t}}=&\operatorname{Sigmoid}({{\mathbf{W}}_{\rm fx}}{{\bm{x}}_{t-1}}+{{\mathbf{W}}_{\rm fu}}{{\bm{u}}_{t}}+{{\mathbf{W}}_{\rm fy}}{{\bm{y}}_{t-1}}+{{\bm{b}}_{\rm f}}),\\
    {\bm{g}^{\rm i}_{t}}=&\operatorname{Sigmoid}({{\mathbf{W}}_{\rm ix}}{{\bm{x}}_{t-1}}+{{\mathbf{W}}_{\rm iu}}{{\bm{u}}_{t}}+{{\mathbf{W}}_{\rm iy}}{{\bm{y}}_{t-1}}+{{\bm{b}}_{\rm i}}),\\
    {\bm{g}^{\rm o}_{t}}=&\operatorname{Sigmoid}({{\mathbf{W}}_{\rm ox}}{{\bm{x}}_{t-1}}+{{\mathbf{W}}_{\rm ou}}{{\bm{u}}_{t}}+{{\mathbf{W}}_{\rm oy}}{{\bm{y}}_{t-1}}+{{\bm{b}}_{\rm o}}),\\
    \end{split}
\end{equation}
where all gates employ the sigmoid function to produce normalized outputs within $[0,1]$. The input gate $\bm{g}^{\rm i}_{t}$ and forget gate $\bm{g}^{\rm f}_{t}$ in \eqref{eq:LSTM_gates} respectively control new information integration and historical state retention in cell state $\bm{c}_t$, serving as a memory buffer for long-term information storage:
\begin{equation}\label{eq:LSTM_hidden}
    \begin{split}
    {{\bm{c}}_{t}}=&\operatorname{Tanh}({{\mathbf{W}}_{\rm cx}}{{\bm{x}}_{t-1}}+{{\mathbf{W}}_{\rm cu}}{{\bm{u}}_{t}}\\&\quad\quad\quad~+{{\mathbf{W}}_{\rm cy}}{{\bm{y}}_{t-1}}+{{\bm{b}}_{\rm c}})\odot{\bm{g}^{\rm i}_{t}}+{{\bm{c}}_{t-1}}\odot{\bm{g}^{\rm f}_{t}},\\
    {{\bm{x}}_{t}}=&\operatorname{Tanh}({{\bm{c}}_{t}})\odot {\bm{g}^{\rm o}_{t}},\\
    {{\hat{\bm{y}}}_{t}}=&{{\mathbf{W}}_{\rm yx}}{{\bm{x}}_{t}}+{{\bm{b}}_{\rm y}}.
    \end{split}
\end{equation}
As defined in \eqref{eq:LSTM_gates}, the output gate $\bm{g}^{\rm o}_{t}$ selectively regulates the exposure of $\bm{c}_t$ to the hidden state $\bm{x}_t$ in \eqref{eq:LSTM_hidden} through the element-wise product $\odot$ between two vectors.

Using a simpler gating control mechanism, GRU offers another popular variant of RNN and achieves comparable performance with LSTM \cite{cho2014properties}. There are only two gating units in the GRU structure:
\begin{equation}\label{eq:GRU_gates}
    \begin{split}
        \bm{g}^{\rm r}_{t} &= \operatorname{Sigmoid}({{\mathbf{W}}_{\rm rx}}{{\bm{x}}_{t-1}}+{{\mathbf{W}}_{\rm ru}}{{\bm{u}}_{t}}+{{\mathbf{W}}_{\rm ry}}{{\bm{y}}_{t-1}}+{{\bm{b}}_{\rm r}}),\\
        \bm{g}^{\rm u}_{t} &= \operatorname{Sigmoid}({{\mathbf{W}}_{\rm ux}}{{\bm{x}}_{t-1}}+{{\mathbf{W}}_{\rm uu}}{{\bm{u}}_{t}}+{{\mathbf{W}}_{\rm uy}}{{\bm{y}}_{t-1}}+{{\bm{b}}_{\rm u}}),
    \end{split} 
\end{equation}
where the update gate $\bm{g}^{\rm u}_{t}$ combines the functionality of $\bm{g}^{\rm i}_{t}$ and $\bm{g}^{\rm f}_{t}$ in \eqref{eq:LSTM_gates}, and the reset gate $\bm{g}^{\rm r}_{t}$ determines the integration ratio of historical information when computing the candidate hidden state $\bm x_t^{'}$. The evolution of hidden states $\bm{x}_t$ follows:
\begin{equation}\label{eq:GRU_hidden}
    \begin{split}
        \bm{x}'_{t} &= \operatorname{Tanh}({{\mathbf{W}}_{\rm xx}}{{\bm{x}}_{t-1}}\odot {\bm{g}^{\rm r}_{t}}+{{\mathbf{W}}_{\rm xu}}{{\bm{u}}_{t}}+{{\mathbf{W}}_{\rm xy}}{{\bm{y}}_{t-1}}+{{\bm{b}}_{\rm x}}),\\
        {{\bm{x}}_{t}} &= {{\bm{x}}_{t-1}}\odot (1-\bm{g}^{\rm u}_{t})+\bm{x}'_{t}\odot {\bm{g}^{\rm u}_{t}},\\
        {{\hat{\bm{y}}}_{t}}&={{\mathbf{W}}_{\rm yx}}{{\bm{x}}_{t}}+{{\bm{b}}_{\rm y}},\\
    \end{split} 
\end{equation}
where $\bm{g}^{\rm u}_{t}$ from \eqref{eq:GRU_gates} determines the proportion of information offered by $\bm{x}_{t-1}$ and $\bm x'_t$.

The proposed innovation-driven structure is readily extensible to LSTM, GRU and other variants of RNN. For LSTM and GRU, we incorporate innovations into both gating units and recurrent hidden units. This naturally gives rise to IGRU and ILSTM, which incorporate innovations into the updates of $\bm{g}^{\rm r}_t$, $\bm{g}^{\rm u}_t$, and $\bm{x}_t^{'}$ in GRU, as well as $\bm{g}^{\rm f}_t$, $\bm{g}^{\rm i}_t$, $\bm{g}^{\rm o}_t$, and $\bm{c}_t$ in LSTM. The mathematical formulations of IGRU and ILSTM are detailed in Table \ref{tab:IRNN_equations} with the single-layer IRNN \eqref{eq:IRNN_1} for comparison. In a similar spirit to \eqref{eq:IRNN_1}, the multi-step ahead prediction based on IGRU and ILSTM can be achieved by simply setting future innovations to zero and replacing $\bm{y}_t$ with ${{\hat{\bm{y}}}_{t}}$.

\begin{table*}[htbp]
\centering
\renewcommand{\arraystretch}{1.5}
\caption{Summary of Innovation-Driven Networks (IRNN, IGRU, and ILSTM) for Time-Series Modeling and Prediction} 
\label{tab:IRNN_equations}
\begin{tabular}{@{}l l@{}}
\toprule
\multicolumn{2}{l}{\textbf{IRNN (Innovation-driven Recurrent Neural Network)}} \\
Hidden state update & ${{\bm{x}}_{t}} =\sigma({{\mathbf{W}}_{\rm xx}}{{\bm{x}}_{t-1}}+{{\mathbf{W}}_{\rm xu}}{{\bm{u}}_{t}}+{{\mathbf{W}}_{\rm xy}}{{\bm{y}}_{t-1}}+{{\mathbf{W}}_{\rm xe}}{{\bm{e}}_{t-1}}+{{\bm{b}}_{\rm x}})$\\
Parameter initialization & ${{\bm{x}}_{0}} =\mathbf{0}, \quad{{\bm{y}}_{0}}=\mathbf{0}, \quad{{\bm{e}}_{0}} =\mathbf{0}$\\
Prediction and innovation & ${{\hat{\bm{y}}}_{t}} = {{\mathbf{W}}_{\rm yx}}{{\bm{x}}_{t}}+{{\bm{b}}_{\rm y}}, \quad {{\bm{e}}_{t}} = {{\bm{y}}_{t}}-{{\hat{\bm{y}}}_{t}}$\\

\midrule
\multicolumn{2}{l}{\textbf{IGRU (Innovation-driven Gated Recurrent Unit)}}\\
Reset gate & $\bm{g}^{\rm r}_{t} = \operatorname{Sigmoid}({{\mathbf{W}}_{\rm rx}}{{\bm{x}}_{t-1}}+{{\mathbf{W}}_{\rm ru}}{{\bm{u}}_{t}}+{{\mathbf{W}}_{\rm ry}}{{\bm{y}}_{t-1}}+{{\mathbf{W}}_{\rm re}}{{\bm{e}}_{t-1}}+{{\bm{b}}_{\rm r}})$ \\
Update gate & $\bm{g}^{\rm u}_{t} = \operatorname{Sigmoid}({{\mathbf{W}}_{\rm ux}}{{\bm{x}}_{t-1}}+{{\mathbf{W}}_{\rm uu}}{{\bm{u}}_{t}}+{{\mathbf{W}}_{\rm uy}}{{\bm{y}}_{t-1}}+{{\mathbf{W}}_{\rm ue}}{{\bm{e}}_{t-1}}+{{\bm{b}}_{\rm u}})$\\
Candidate hidden state & $\bm{x}'_{t} = \operatorname{Tanh}({{\mathbf{W}}_{\rm xx}}\cdot {{\bm{x}}_{t-1}}\odot {\bm{g}^{\rm r}_{t}}+{{\mathbf{W}}_{\rm xu}}{{\bm{u}}_{t}}+{{\mathbf{W}}_{\rm xy}}{{\bm{y}}_{t-1}}+{{\mathbf{W}}_{\rm xe}}{{\bm{e}}_{t-1}}+{{\bm{b}}_{\rm x}})$\\
Hidden state update & ${{\bm{x}}_{t}} = {{\bm{x}}_{t-1}}\odot (1-\bm{g}^{\rm u}_{t})+\bm{x}'_{t}\odot {\bm{g}^{\rm u}_{t}}$\\
Parameter initialization & ${{\bm{x}}_{0}}=\bm{0}, \quad{{\bm{y}}_{0}}=\bm{0},\quad {{\bm{e}}_{0}}=\bm{0}$\\
Prediction and innovation & ${{\hat{\bm{y}}}_{t}}={{\mathbf{W}}_{\rm yx}}{{\bm{x}}_{t}}+{{\bm{b}}_{\rm y}}, \quad {{\bm{e}}_{t}}={{\bm{y}}_{t}}-{{\hat{\bm{y}}}_{t}}$\\

\midrule
\multicolumn{2}{l}{\textbf{ILSTM (Innovation-driven Long Short-Term Memory)}}\\
Forget gate & ${\bm{g}^{\rm f}_{t}}=\operatorname{Sigmoid}({{\mathbf{W}}_{\rm fx}}{{\bm{x}}_{t-1}}+{{\mathbf{W}}_{\rm fu}}{{\bm{u}}_{t}}+{{\mathbf{W}}_{\rm fy}}{{\bm{y}}_{t-1}}+{{\mathbf{W}}_{\rm fe}}{{\bm{e}}_{t-1}}+{{\bm{b}}_{\rm f}})$\\
Input gate & ${\bm{g}^{\rm i}_{t}}=\operatorname{Sigmoid}({{\mathbf{W}}_{\rm ix}}{{\bm{x}}_{t-1}}+{{\mathbf{W}}_{\rm iu}}{{\bm{u}}_{t}}+{{\mathbf{W}}_{\rm iy}}{{\bm{y}}_{t-1}}+{{\mathbf{W}}_{\rm ie}}{{\bm{e}}_{t-1}}+{{\bm{b}}_{\rm i}})$\\
Output gate & ${\bm{g}^{\rm o}_{t}}=\operatorname{Sigmoid}({{\mathbf{W}}_{\rm ox}}{{\bm{x}}_{t-1}}+{{\mathbf{W}}_{\rm ou}}{{\bm{u}}_{t}}+{{\mathbf{W}}_{\rm oy}}{{\bm{y}}_{t-1}}+{{\mathbf{W}}_{\rm oe}}{{\bm{e}}_{t-1}}+{{\bm{b}}_{\rm o}})$\\
Cell state & ${{\bm{c}}_{t}}=\operatorname{Tanh}({{\mathbf{W}}_{\rm cx}}{{\bm{x}}_{t-1}}+{{\mathbf{W}}_{\rm cu}}{{\bm{u}}_{t}}+{{\mathbf{W}}_{\rm cy}}{{\bm{y}}_{t-1}}+{{\mathbf{W}}_{\rm ce}}{{\bm{e}}_{t-1}}+{{\bm{b}}_{\rm c}})\odot{\bm{g}^{\rm i}_{t}}+{{\bm{c}}_{t-1}}\odot{\bm{g}^{\rm f}_{t}}$\\
Hidden state update & ${{\bm{x}}_{t}}=\operatorname{Tanh}({{\bm{c}}_{t}})\odot {\bm{g}^{\rm o}_{t}}$\\
Parameter initialization & ${{\bm{x}}_{0}}=\bm{0},\quad{{\bm{y}}_{0}}=\bm{0},\quad{{\bm{e}}_{0}}=\bm{0}$\\
Prediction and innovation & ${{\hat{\bm{y}}}_{t}}={{\mathbf{W}}_{\rm yx}}{{\bm{x}}_{t}}+{{\bm{b}}_{\rm y}}, \quad {{\bm{e}}_{t}}={{\bm{y}}_{t}}-{{\hat{\bm{y}}}_{t}}$\\

\bottomrule
\end{tabular}
\end{table*}

\subsection{A Customized Training Scheme for IRNN}
In this subsection, we present a tailored scheme for training IRNNs. Before proceeding, we first recall BPTT, a prevalent algorithm for training conventional RNNs and their variants \cite{werbos1990backpropagation}. Given a time-series trajectory $\mathcal{T} = \{ (\bm{u}_t,\bm{y}_t)\}_{t=1}^{T}$ and parameters $\Theta$ of RNN, a loss function $\ell(\mathcal{T}; \Theta)$ can be defined to drive the optimization of $\Theta$ via gradient descent:
\begin{equation}
\label{eq:GD}
    \Theta_{k+1} = \Theta_k - \eta \left. \frac{\partial \ell ( \mathcal{T}; \Theta )}{\partial \Theta}\right|_{\Theta = \Theta_k},
\end{equation}
where $\Theta_{k}$ represents values of network parameters at the $k$th iteration, and $\eta > 0$ is the learning rate. In \eqref{eq:GD}, the effective computation of the gradient term for recurrent architectures is enabled by BPTT. The mean squared error (MSE) has been widely adopted to evaluate the overall accuracy of multi-step prediction:
\begin{equation}\label{eq:MSE segment}
    \ell(\mathcal{T}; \Theta)=\frac{1}{T_{\rm f}}\sum_{t=1}^{T_{\rm f}}\left \| \bm{y}_{t}-\hat{\bm{y}}_{t} \right \|_2^2 ,
\end{equation}
where $\left \| \cdot\right\|_2$ is the $\ell_2$-norm, and $T_{\rm f}$ is the length of future prediction horizon. $\hat{\bm y}_t$ is given by \eqref{eq:IRNN_2} after the network learns system dynamics from the past time-series trajectory with length $T_{\rm p}$.

However, the updating rule \eqref{eq:GD} does not apply straightforwardly to IRNNs \eqref{eq:generalIRNN} due to input augmentation based on innovations. More formally, a single data trajectory for training IRNN can be expressed as $\mathcal{T}(\Theta) = \{ (\bm{u}_t,\bm{y}_t, \bm{e}_t(\Theta) )\}_{t=1}^{T}$, where the realization of $\bm{e}_t (\Theta)$ inherently depends on network parameters $\Theta$. Unfortunately, the standard BPTT algorithm does not take such dependence into account, which is a key issue in training IRNN.

\begin{figure}[htbp]
    \centering
    \includegraphics[width=\linewidth]{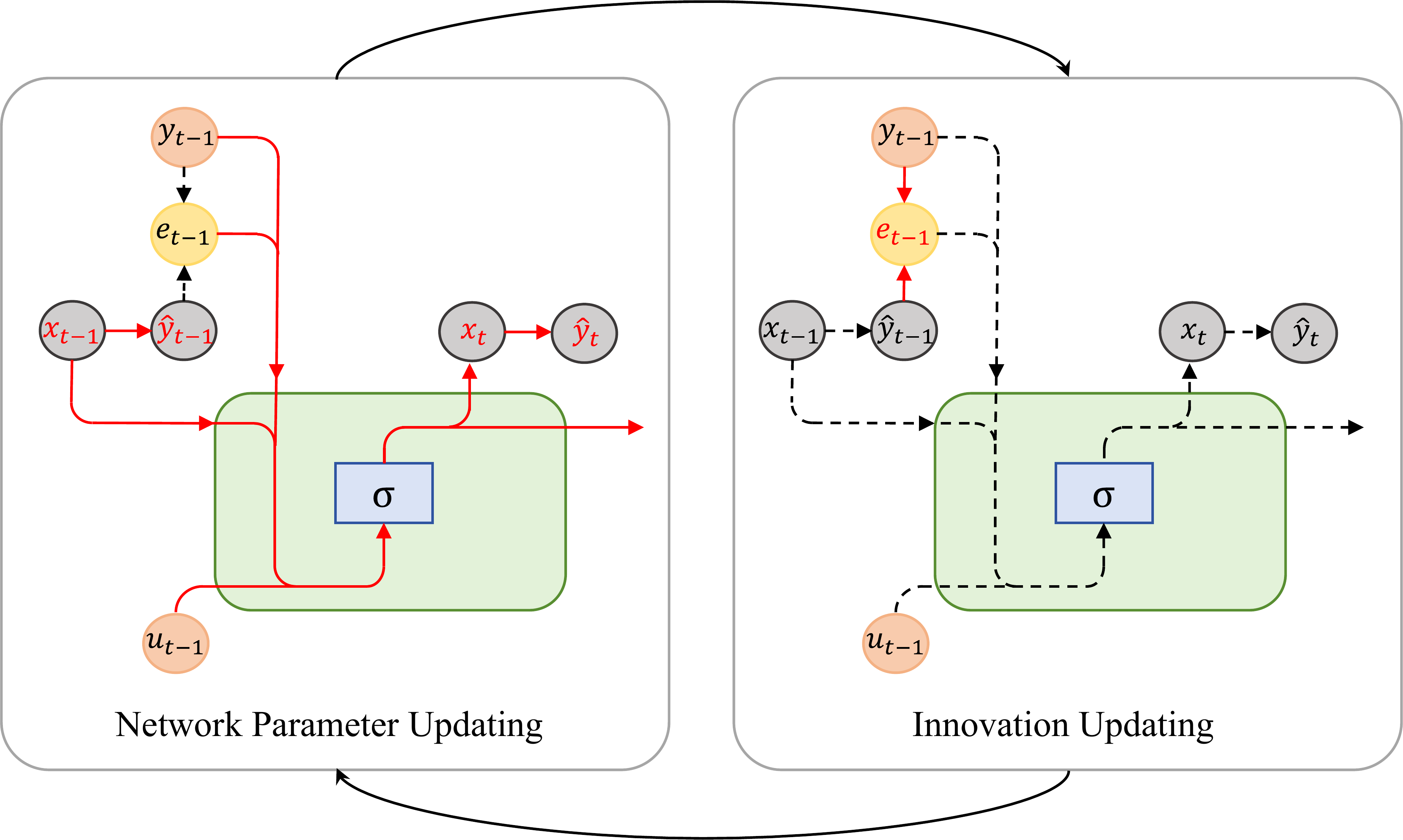}
    \caption{IU-BPTT for training single-layer IRNN. The updating of network parameters (left) and the updating of innovations (right) alternate in a cyclic manner. Dashed and red solid arrows indicate inactive and active data flow, respectively.}
    \label{fig:IU}
\end{figure}

To address the coupling between inputs and network parameters, we propose IU-BPTT, an efficient modification of BPTT tailored to training IRNN. Fig. \ref{fig:IU} exemplifies the iterative training procedure of IU-BPTT for a single-layer IRNN \eqref{eq:IRNN_1}. It decomposes the optimization process \eqref{eq:GD} into iterations between the following two steps.
\begin{itemize}
    \item \textbf{Network parameter updating}: Within each training epoch, we propose to fix the data trajectory based on current values of network parameters, which enables to compute the gradient by invoking the generic BPTT algorithm. The policy of gradient descent can be written as:
\begin{equation}
    \label{eq:IU-GD}
    \Theta_{k+1} = \Theta_k - \eta \left. \frac{\partial \ell  \left( \mathcal{T}(\Theta_k); \Theta \right)}{\partial \Theta} \right|_{\Theta = \Theta_k}.
\end{equation}

    \item \textbf{Innovation updating}: After performing gradient descent in an epoch, we update the innovations $\{\bm{e}_t\}_{t=1}^T$ based on the latest network parameters $\Theta_{k}$ at hand:
\begin{equation}
    \label{eq:inno_updating}
    \bm{e}_t (\Theta_{k}) = \bm{y}_t-\hat{\bm{y}}_t,
\end{equation}
where $\hat{\bm{y}}_t$ denotes the resultant one-step prediction of $\bm{y}_t$.
\end{itemize}

%\subsubsection{Training Algorithm and Details}
Algorithm \ref{alg:IRNN train} outlines the complete implementation procedure of our IU-BPTT for training IRNNs. An obvious appeal is that off-the-shelf routines of BPTT can be invoked to facilitate the algorithmic implementation. Besides, two strategies are devised to further improve the training efficiency.

\begin{algorithm}[hbtp]
\caption{Training IRNN with IU-BPTT}\label{alg:IRNN train}
\begin{algorithmic}[0] 
\STATE \textbf{INPUT:} Training set $\mathcal{D}_\text{T}$.
\STATE \textbf{Parameters:} Learning rate $\eta \in \mathbb{R}^{+}$, number of epochs $n \in \mathbb{N}^{+}$, the innovation updating interval $N$, length of segmented time series trajectory $T$, length of the past trajectory $T_{\rm p}$, length of future prediction horizon $T_{\rm f}$.
\end{algorithmic}
\begin{algorithmic}[1]
\STATE \textbf{Initialize} model parameters $\Theta$
\STATE \textbf{for} epoch $k = 1$ \textbf{to} $n$ \textbf{do}
\STATE \hspace{0.5cm} \textit{\textbf{Parameter Updating:}}
\STATE \hspace{0.5cm} \textbf{for} each trajectory $\mathcal{T}=\{({\bm{u}_t},\bm{y}_t, \bm{e}_t)\}^T_{t=1}$ \textbf{in} $\mathcal{D}_\text{T}$
\STATE \hspace{1cm} \textbf{Initialize} $\bm{x}_0$, $\hat{\bm{y}}_0 = 0$
\STATE \hspace{1cm} \textbf{for} $t = 1$ \textbf{to} $T_{\rm p}$ \textbf{do}
\STATE \hspace{1.5cm} Update $\bm{x}_t$, $\hat{\bm{y}}_{t}$ with fixed $\Theta$ by \eqref{eq:IRNN_1} 
\STATE \hspace{1cm} \textbf{end for}
\STATE \hspace{1cm} \textbf{for} $t=T_{\rm p} + 1$ \textbf{to} $T$ \textbf{do}
\STATE \hspace{1.5cm} Update $\bm{x}_t$, $\hat{\bm{y}}_{t}$ with fixed $\Theta$ by \eqref{eq:IRNN_2}
\STATE \hspace{1cm} \textbf{end for}
\STATE \hspace{1cm} Compute $\ell(\mathcal{T}; \Theta)$ by \eqref{eq:MSE segment}
\STATE \hspace{2cm} Update $\Theta$ via IU-BPTT by \eqref{eq:IU-GD}
\STATE \hspace{0.5cm} \textbf{end for}

\STATE \hspace{0.5cm} \textit{\textbf{Innovation Updating:}}
\STATE \hspace{0.5cm} \textbf{if} $k ~\operatorname{mod} ~N = 0$ \textbf{then}
\STATE \hspace{1cm} \textbf{for} each $\mathcal{T}=\{({\bm{u}_t},\bm{y}_t, \bm{e}_t)\}^T_{t=1}$ \textbf{in} $\mathcal{D}_\text{T}$
\STATE \hspace{1.5cm} \textbf{for} $t = 1$ \textbf{to} $T_{\rm p}$ \textbf{do}
\STATE \hspace{2cm} Compute $\hat{\bm{y}}$ with updated $\Theta$ by \eqref{eq:IRNN_1}
\STATE \hspace{2cm} Update $\bm{e}_t(\mathcal{T}; \Theta)$ by \eqref{eq:inno_updating}
\STATE \hspace{1.5cm} \textbf{end for}
\STATE \hspace{1cm} \textbf{end for}
\STATE \hspace{0.5cm} \textbf{end if}
\STATE \textbf{end for}
\end{algorithmic}
\end{algorithm}

\begin{itemize}
    \item \textbf{Trade-off in innovation updating}: In IU-BPTT, the updating of $\bm{e}_t$ in each epoch inevitably leads to some extra computations. One can further achieve a flexible compromise between computational costs and fidelity of gradient computations by performing innovation updating per $N(>1)$ consecutive epochs.

    \item \textbf{Dataset segmentation for parallelized training}: For a data trajectory with length $T$, the computational complexity of BPTT is known to be $\mathcal{O}(T^2)$, which becomes prohibitively expensive when facing long sequences \cite{chauvin2013backpropagation}. A useful and ubiquitous remedy is to split a long sequence into multiple shorter segments, which not only lowers the complexity but also allows for parallelized training \cite{750549, lin2023segrnn}. An illustration is provided in Fig. \ref{fig:dataset}, where multiple trajectories are produced by splitting a single long trajectory, and each trajectory is then augmented with its own innovations to implement IU-BPTT.
    
    \begin{figure}[hbtp]
    \centering
    \includegraphics[width=0.9\linewidth]{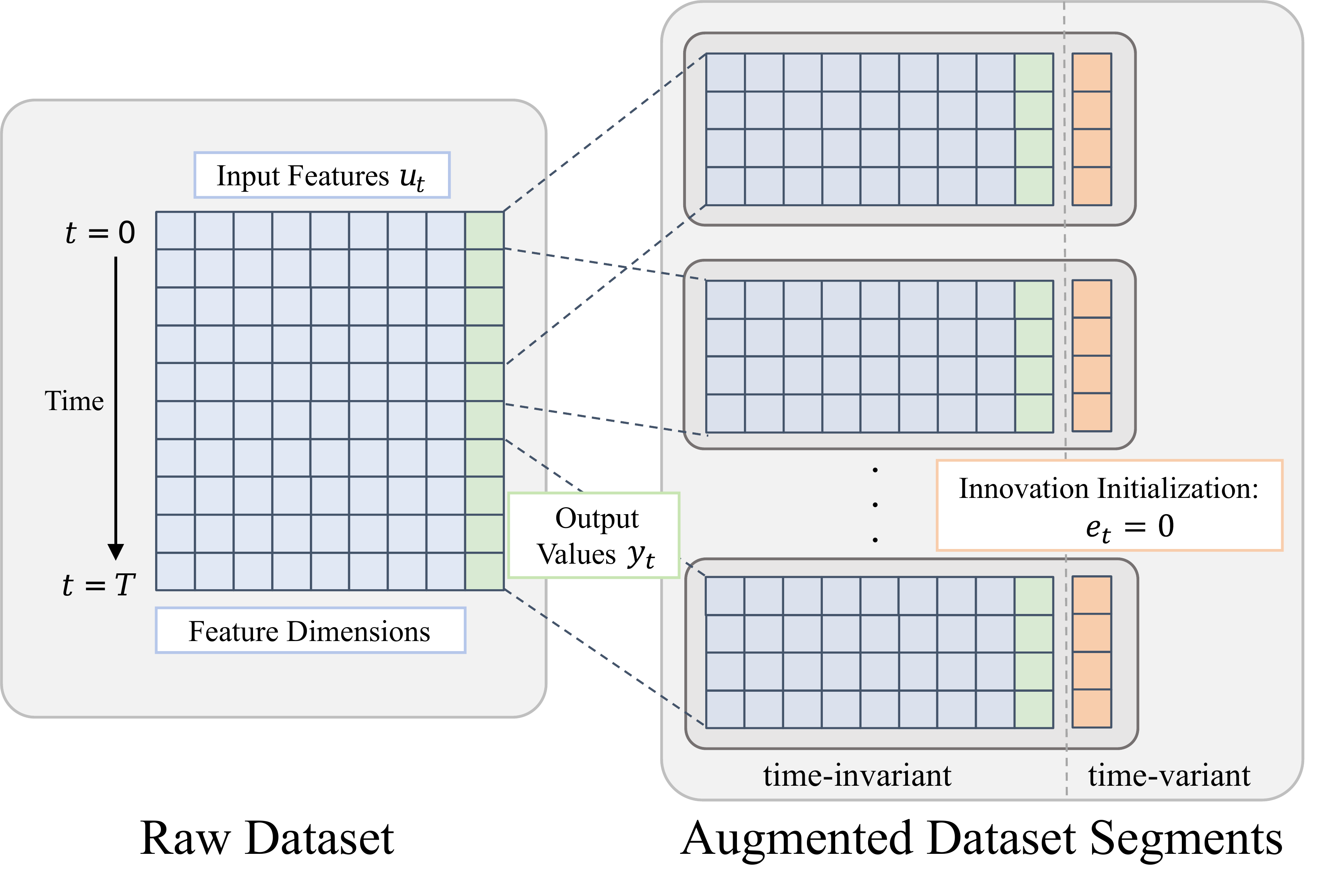}
    \caption{Dataset segmentation and augmentation. The raw time-series trajectory with a length of $T$ (left) is separated into multiple shorter segments (right). Each segment is augmented with innovations $\bm{e}_t$, which are initialized to zero when training is initiated.}
    \label{fig:dataset}
    \end{figure}
    
\end{itemize}

\section{Experimental Studies}
\label{sec:expriment}
In this section, we conduct a comparative analysis between single-layer IRNN, IGRU and ILSTM, and their RNN counterparts without innovation on real-world time-series datasets. 

\subsection{Settings}

\subsubsection{Data description}
We use the ETT (Electricity Transformer Temperature) database \cite{zhou2021informer} to evaluate the performance of IRNN, IGRU, and ILSTM outlined in Table \ref{tab:IRNN_equations}. As shown in Table \ref{tab:dataset}, ETT comprises four datasets, namely ETTh1, ETTh2, ETTm1, and ETTm2, which contain electricity transformer data collected between July 2016 to July 2018 from two different counties in China. 

\begin{table}[htbp]
    \centering
    \caption{Sampling Details and Feature Descriptions of ETT}
    \begin{tabular}{c|cccc}
    \toprule
    Dataset & ETTh1 & ETTh2 & ETTm1 & ETTm2 \\
    \midrule
    Sampling Interval & 1 hour & 1 hour & 15 minutes & 15 minutes\\
    Sequential Length & 17420 & 17420 & 69680 & 69680\\
    Input Feature & \multicolumn{4}{c}{Power Loads $\times ~6$} \\
    Output Features & \multicolumn{4}{c}{Oil Temperature $\times ~1$} \\
    \bottomrule
    \end{tabular}
    \label{tab:dataset}
\end{table}

ETT provides a variety of time-series data with different trends, dispersion, and time intervals, serving as a widely used benchmark in time-series prediction \cite{zhou2022fedformer, qin2024hierarchically}. ETT comprises 7 features: the oil temperature and 6 types of power loads, where oil temperature is commonly treated as the prediction target $\bm{y}_t$, and the remaining 6 features are served as inputs $\bm{u}_t$ to model the underlying system dynamics \cite{wang2024deep, liang2024foundation}.

\subsubsection{Implementation details}
\label{sec:implementation}
We implement three mainstream recurrent networks, RNN, GRU, and LSTM with innovation in the following experiments, namely IRNN, IGRU, and ILSTM. Table \ref{tab:settings} lists the key hyperparameter settings for training these RNNs and their innovation-driven counterparts. Considering the periodicity of ETT, we partition the raw time series trajectory into segments with length $T=29$, where the history trajectory length is fixed at $T_{\rm p}=24$ for feature learning, corresponding to 24-hour periods in ETTh1/ETTh2 and 6-hour periods in ETTm1/ETTm2. The remaining portion, with length $T_{\rm f}=5$, is used for multi-step prediction. The segmented datasets are shuffled and split into training, validation and test sets with a ratio of $6:2:2$. All networks are optimized with the ADAM optimizer \cite{kingma2014adam}, and early stopping is triggered once the validation loss does not decrease in 5 consecutive epochs. All networks are implemented in PyTorch \cite{paszke2019pytorch} and trained 20 times on a single NVIDIA GeForce RTX 4070 GPU.

\begin{table}[htbp]
    \centering
    \caption{Hyperparameter Settings in Training IRNNs}
    \begin{tabular}{ccc}
    \toprule
    Hyperparameters & (I)RNN & (I)GRU \& (I)LSTM \\
    \midrule
    Learning rate $\eta$ & $6\times10^{-4}$ & $3\times10^{-4}$ \\
    Length of past trajectory $T_{\rm p}$ & $24$ & $24$ \\
    Length of prediction $T_{\rm f}$ & $5$ & $5$ \\
    Batch size & $64$ & $64$ \\
    Size of hidden state & $128$ & $128$ \\
    Number of hidden layer & $1$ & $1$ \\
    Early stop tolerance $n_{\rm tol}$ & $5$ & $5$ \\
    Number of epochs $n$ & $100$ & $100$ \\
    Innovation updating interval $N$ & $1$ & $1$ \\
    \bottomrule
    \end{tabular}
    \label{tab:settings}
\end{table}

\subsection{Prediction Performance}
\label{sec:prediction accuracy}
\subsubsection{Improvements in prediction accuracy}
To assess prediction accuracy, six network architectures are evaluated on four datasets and their test errors are summarized in Table \ref{tab:result_ETT}. 
Because of the usage of innovations as extra inputs, the single-layer IRNN, IGRU, and ILSTM show better multi-step prediction performance than the single-layer RNN, GRU, and LSTM on four test datasets, where ILSTM achieves the greatest improvement, with \textbf{31\%} (0.0276 $\to$ 0.0190) MSE reduction on ETTh1, \textbf{31\%} (0.0326 $\to$ 0.0212) on ETTh2, \textbf{19\%} (0.0094 $\to$ 0.0076) on ETTm1, and \textbf{27\%} (0.0036 $\to$ 0.0026) on ETTm2. IRNN and IGRU also achieve obvious MSE reductions of \textbf{15\%} (0.0317$\to$0.0269) and \textbf{22\%} (0.0359$\to$0.0280) on average across four datasets, respectively. We also report the prediction performance of Naïve Prediction in Table \ref{tab:result_ETT} as a benchmark. Its prediction at time instance $t+1$ equals to the observation at the previous instance $t$, i.e., $\hat{\bm{y}}_{t+1}=\bm{y}_t$ \cite{box2015time}. If a particular network does perform as well as Naïve Prediction, then the network is considered to not desirably capture dynamics in the observed data. It can be seen from Table \ref{tab:result_ETT} that the single-layer RNN, GRU, and LSTM fail to outperform Naïve Prediction in terms of 1-step prediction on ETTh1 and ETTm1. By contrast, the 1-step prediction performance of three innovation-driven RNNs, including the single-layer IRNN, IGRU, and ILSTM, is better than that of Naïve Prediction on ETTh1, and comparable with that of Naïve Prediction on ETTm1. This highlights the effectiveness of introducing innovations in improving the prediction performance of RNN.

\begin{table*}[hbtp]
  \centering
  \begin{threeparttable}
  \caption{Test Error (MSE) on ETT Datasets}
  \label{tab:result_ETT}
    \begin{tabularx}{0.8\textwidth}{c|c|XXXXXX}
    \toprule
    Dataset & MSE Loss & 1-step & 2-step & 3-step & 4-step & 5-step & Average\\
    \specialrule{0.8pt}{1pt}{1pt}
    \multirow{10}{*}{ETTh1} 
    & Single-layer RNN  & 0.0135 & 0.0197 & 0.0279 & 0.0351 & 0.0400 & 0.0272\\
    & Single-layer IRNN & \textbf{0.0101} & \textbf{0.0188} & \textbf{0.0265} & \textbf{0.0338} & \textbf{0.0386} & \textbf{0.0255}\\
    & Improvement  & 25.71\%  & 4.70\%   & 4.90\%   & 3.90\%   & 3.60\%   & 6.30\%\\
    \cmidrule(lr){2-8}
    \multirow{3}{*}{} 
    & GRU          & 0.0120 & 0.0215 & 0.0301 & 0.0382 & 0.0437 & 0.0291\\
    & IGRU         & \textbf{0.0103} & \textbf{0.0197} & \textbf{0.0281} & \textbf{0.0358} & \textbf{0.0414} & \textbf{0.0271}\\
    & Improvement  & 14.10\%  & 8.55\%   & 6.49\%   & 6.08\%   & 5.41\%   & 6.99\%\\
    \cmidrule(lr){2-8}
    \multirow{3}{*}{} 
    & LSTM          & 0.0163 & 0.0215 & 0.0275 & 0.0338 & 0.0386 & 0.0276\\
    & ILSTM         & \textbf{0.0091} & \textbf{0.0149} & \textbf{0.0203} & \textbf{0.0239} & \textbf{0.0267} & \textbf{0.0190}\\
    & Improvement  & 44.00\%  & 30.63\%  & 26.12\%  & 29.26\%  & 30.84\%  & 31.03\%\\
    \cmidrule(lr){2-8}
    \multirow{1}{*}{} 
    & Naïve        & 0.0107 & 0.0225 & 0.0344 & 0.0476 & 0.0575 & 0.0346\\
    \specialrule{0.8pt}{1pt}{1pt}
    
    \multirow{10}{*}{ETTh2}
    & Single-layer RNN  & 0.0082 & 0.0148 & 0.0302 & 0.0453 & 0.0603 & 0.0317\\
    & Single-layer IRNN & \textbf{0.0041} & \textbf{0.0125} & \textbf{0.0255} & \textbf{0.0393} & \textbf{0.0531} & \textbf{0.0269}\\
    & Improvement  & 49.30\%  & 15.22\% & 15.55\% & 13.35\% & 11.80\% & 15.20\%\\
    \cmidrule(lr){2-8}
    \multirow{3}{*}{}
    & GRU          & 0.0043 & 0.0147 & 0.0309 & 0.0523 & 0.0764 & 0.0359\\
    & IGRU         & \textbf{0.0037} & \textbf{0.0122} & \textbf{0.0257} & \textbf{0.0404} & \textbf{0.0578} & \textbf{0.0280}\\
    & Improvement  & 13.58\% & 17.13\% & 19.64\% & 22.71\% & 24.36\% & 22.19\%\\
    \cmidrule(lr){2-8}
    \multirow{3}{*}{}
    & LSTM         & 0.0083 & 0.0168 & 0.0300 & 0.0454 & 0.0624 & 0.0326\\
    & ILSTM        & \textbf{0.0033} & \textbf{0.0101} & \textbf{0.0196} & \textbf{0.0305} & \textbf{0.0427} & \textbf{0.0212}\\
    & Improvement  & 60.50\% & 39.78\% & 34.56\% & 32.91\% & 31.58\% & 31.03\%\\
    \cmidrule(lr){2-8}
    \multirow{1}{*}{}
    & Naïve        & 0.0119 & 0.0422 & 0.0886 & 0.1378 & 0.1892 & 0.0936\\
    \specialrule{0.8pt}{1pt}{1pt}

    \multirow{10}{*}{ETTm1}
    & Single-layer RNN  & 0.0030 & 0.0058 & 0.0089 & 0.0121 & 0.0153 & 0.0090\\
    & Single-layer IRNN & \textbf{0.0025} & \textbf{0.0055} & \textbf{0.0082} & \textbf{0.0110} & \textbf{0.0136} & \textbf{0.0082}\\
    & Improvement  & 16.63\%  & 5.69\%  & 7.89\%  & 9.19\%  & 10.76\%  & 9.51\%\\
    \cmidrule(lr){2-8}
    \multirow{3}{*}{}
    & GRU          & 0.0028 & 0.0056 & 0.0084 & 0.0113 & 0.0141 & 0.0084\\
    & IGRU         & \textbf{0.0025} & \textbf{0.0054} & \textbf{0.0081} & \textbf{0.0109} & \textbf{0.0138} & \textbf{0.0081}\\
    & Improvement  & 10.20\% & 3.76\% & 3.18\% & 2.98\% & 2.49\% & 3.44\%\\
    \cmidrule(lr){2-8}
    \multirow{3}{*}{}
    & LSTM         & 0.0049 & 0.0066 & 0.0090 & 0.0117 & 0.0146 & 0.0094\\
    & ILSTM        & \textbf{0.0025} & \textbf{0.0052} & \textbf{0.0077} & \textbf{0.0102} & \textbf{0.0126} & \textbf{0.0076}\\
    & Improvement(\%)  & 50.02\% & 21.00\% & 14.38\% & 13.47\% & 13.74\% & 18.64\%\\
    \cmidrule(lr){2-8}
    \multirow{1}{*}{}
    & Naïve        & 0.0025 & 0.0056 & 0.0085 & 0.0116 & 0.0148 & 0.0086\\
    \specialrule{0.8pt}{1pt}{1pt}
    
    \multirow{10}{*}{ETTm2}
    & Single-layer RNN   & 0.0011 & 0.0016 & 0.0028 & 0.0046 & 0.0063 & 0.0033\\
    & Single-layer IRNN         & \textbf{0.0004} & \textbf{0.0013} & \textbf{0.0020} & \textbf{0.0036} & \textbf{0.0054} & \textbf{0.0025}\\
    & Improvement(\%)  & 66.27\% & 18.11\% & 26.01\% & 20.28\% & 14.45\% & 22.00\%\\
    \cmidrule(lr){2-8}
    \multirow{3}{*}{}
    & GRU          & 0.0006 & 0.0017 & 0.0026 & 0.0044 & 0.0065 & 0.0032\\
    & IGRU         & \textbf{0.0004} & \textbf{0.0014} & \textbf{0.0022} & \textbf{0.0040} & \textbf{0.0060} & \textbf{0.0028}\\
    & Improvement(\%)  & 31.79\% & 15.07\% & 13.44\% & 9.87\% & 7.86\% & 10.97\%\\
    \cmidrule(lr){2-8}
    \multirow{3}{*}{}
    & LSTM         & 0.0013 & 0.0021 & 0.0029 & 0.0048 & 0.0069 & 0.0036\\
    & ILSTM        & \textbf{0.0005} & \textbf{0.0014} & \textbf{0.0021} & \textbf{0.0037} & \textbf{0.0055} & \textbf{0.0026}\\
    & Improvement(\%)  & 62.35\% & 32.82\% & 27.68\% & 23.39\% & 20.33\% & 26.74\%\\
    \cmidrule(lr){2-8}
    \multirow{1}{*}{}
    & Naïve        & 0.0009 & 0.0034 & 0.0064 & 0.0110 & 0.0162 & 0.0076\\
    \bottomrule
  \end{tabularx}
  \begin{tablenotes}
    \footnotesize
    \item Bold fonts highlight the smaller MSE on test set compared to the corresponding networks on the same dataset (e.g., MSE of LSTM and ILSTM on ETTh1).
  \end{tablenotes}
  \end{threeparttable}
\end{table*}

Among all prediction steps, the 1-step prediction performances of IRNNs show the most significant improvement among all prediction steps (see Table \ref{tab:result_ETT}), with MSE reductions ranging from $\textbf{40\%}$ to $\textbf{60\%}$. With the prediction horizon getting longer, the prediction performance improvements of IRNNs become less significant, as the error-correction effect of innovations tends to vanish over time. Despite this, the 5-step prediction of IRNNs is still better than that of their counterparts without innovation, achieving about $\textbf{10\%}$ reductions in test errors.

\subsubsection{Visualization of multi-step prediction}
\label{sec:Prediction Visualization}
\begin{figure*}[hbtp]
    \centering
    \subfloat[]{
        \includegraphics[width=0.5\linewidth]{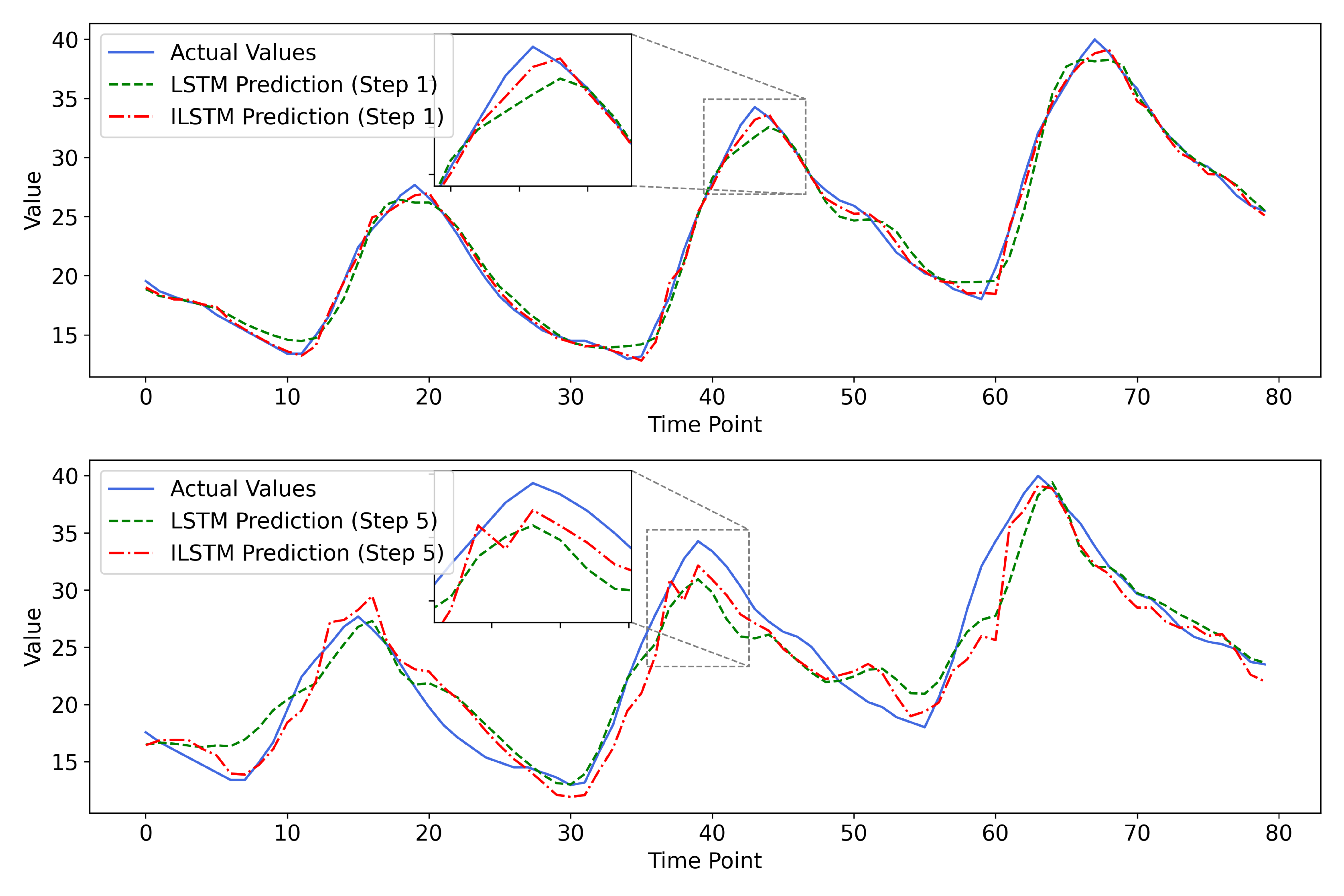}
        \label{fig:ETTh2visual}}
    \subfloat[]{
        \includegraphics[width=0.5\linewidth]{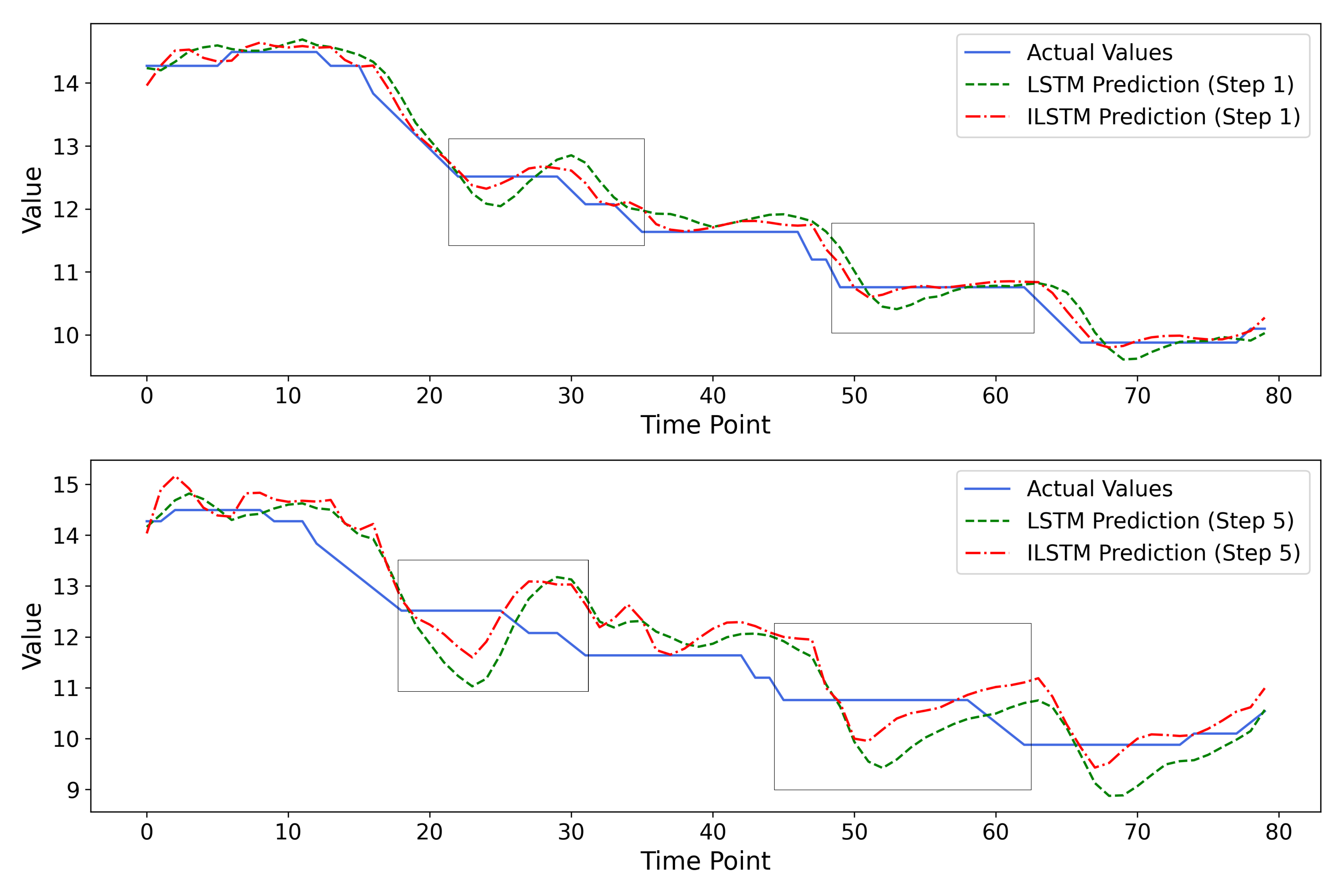}
        \label{fig:ETTm2visual}}
    \caption{Visualization of prediction results. We display the prediction results of LSTM and ILSTM on continuous data with length of 80 in two different time scales. (a) Prediction on ETTh2. (b) Prediction on ETTm2.}
    \label{fig:dataVisual}
\end{figure*}

To give deeper insights into the improved performance of IRNNs, we visualize in Fig. \ref{fig:dataVisual} the 1-step and 5-step predictions of LSTM and ILSTM on two particular data trajectories of length 80. Fig. \ref{fig:ETTh2visual} presents the predictions of a trajectory from ETTh2, which shows periodic characteristics and exhibits an oscillating upward trend. In this case, both LSTM and ILSTM perform well. In terms of 1-step prediction, ILSTM shows better capability in capturing trend variations around turning points than LSTM without innovations. The 5-step predictions of both ILSTM and LSTM are poorer than their 1-step predictions, where ILSTM shows an obvious effect of error correction, especially after a large prediction error appears. This error-correcting capability of ILSTM is more evident when handling the trajectory from ETTh2 in Fig \ref{fig:ETTm2visual}. It can be seen that the predictions of both ILSTM and LSTM fluctuate around the true outputs. ILSTM can correct errors more effectively when there are significant prediction errors, which eventually leads to better prediction accuracy.

\subsubsection{Sizes of network parameterization}
In principle, IRNNs has more parameters than RNNs because of the introduction of innovations. The increased size of network parameterization is determined by the dimension of hidden state $\bm{x}_t$, the dimension of output $\bm{y}_t$, and network modules where innovations are added. A natural question is that how many extra parameters are caused by introducing innovations into RNNs. Thus we present in Table \ref{tab:ParaCompare} the sizes of network parameterization of RNNs and corresponding IRNNs. It can be seen that introducing innovations into the single-layer RNN, GRU, and LSTM resulting in only a $0.73\%$ increase in parameter sizes. Nevertheless, this eventually leads to a $13.25\%$ MSE reduction for RNN and an impressive $26.86\%$ reduction for LSTM. This demonstrates that incorporating innovations into RNN structures only causes a minor increase in parameter sizes, which does not cause any higher requirements on memory for network inference.

\begin{table}[htbp]
    \centering
    \begin{threeparttable}
        \renewcommand{\arraystretch}{1.5}
        \caption{Network Parameter Sizes and MSE Reduction}
        \label{tab:ParaCompare}
        
        \begin{tabular}{c|cc|cc}
        \toprule
        \multirow{2}{*}{Network} & \multicolumn{2}{c|}{Parameter Sizes} & \multirow{2}{*}{\shortstack{Parameter \\ Increase}} & \multirow{2}{*}{\shortstack{MSE \\ Reduction}} \\ \cline{2-3}
         & Original & Innovation & & \\ 
         \midrule
        Single-layer RNN  & 17.5k & 17.7k & 0.73\% & 13.25\% \\ 
        GRU  & 35.1k & 35.3k & 0.73\% & 10.90\% \\ 
        LSTM & 69.8k & 70.3k & 0.73\% & 26.86\% \\
        \bottomrule
        \end{tabular}
        
        \begin{tablenotes}
            \footnotesize
            \item The percentage reduction in MSE is calculated by averaging the experimental results across four datasets in Table \ref{tab:result_ETT}.
        \end{tablenotes}
    \end{threeparttable}
\end{table}

\subsection{Training Performance of IU-BPTT}
\begin{figure*}[hbtp]
    \centering
    \subfloat[]{
        \includegraphics[width=0.33\textwidth]{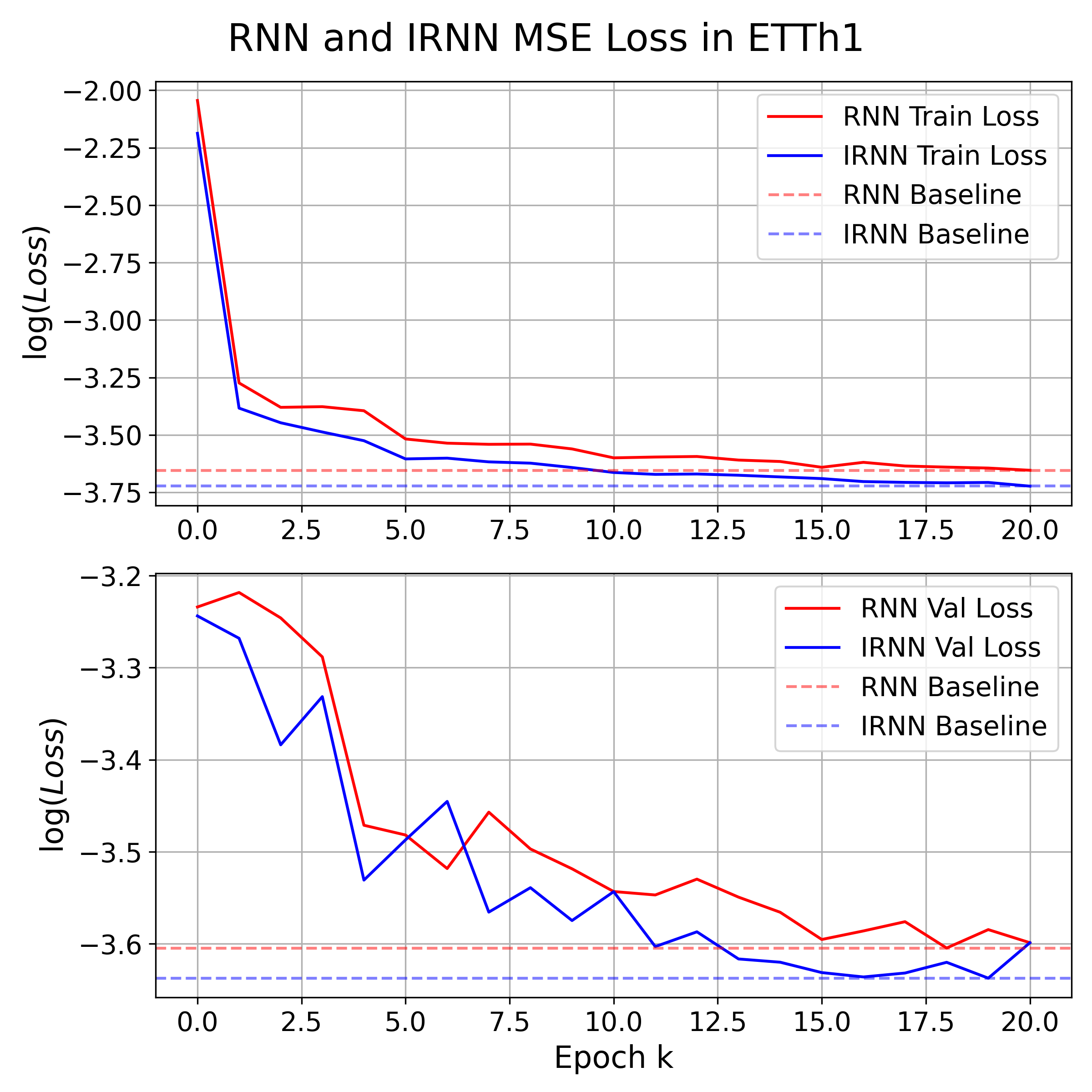}
        \label{fig:loss_RNN_ETTh1}}
    \subfloat[]{
        \includegraphics[width=0.33\textwidth]{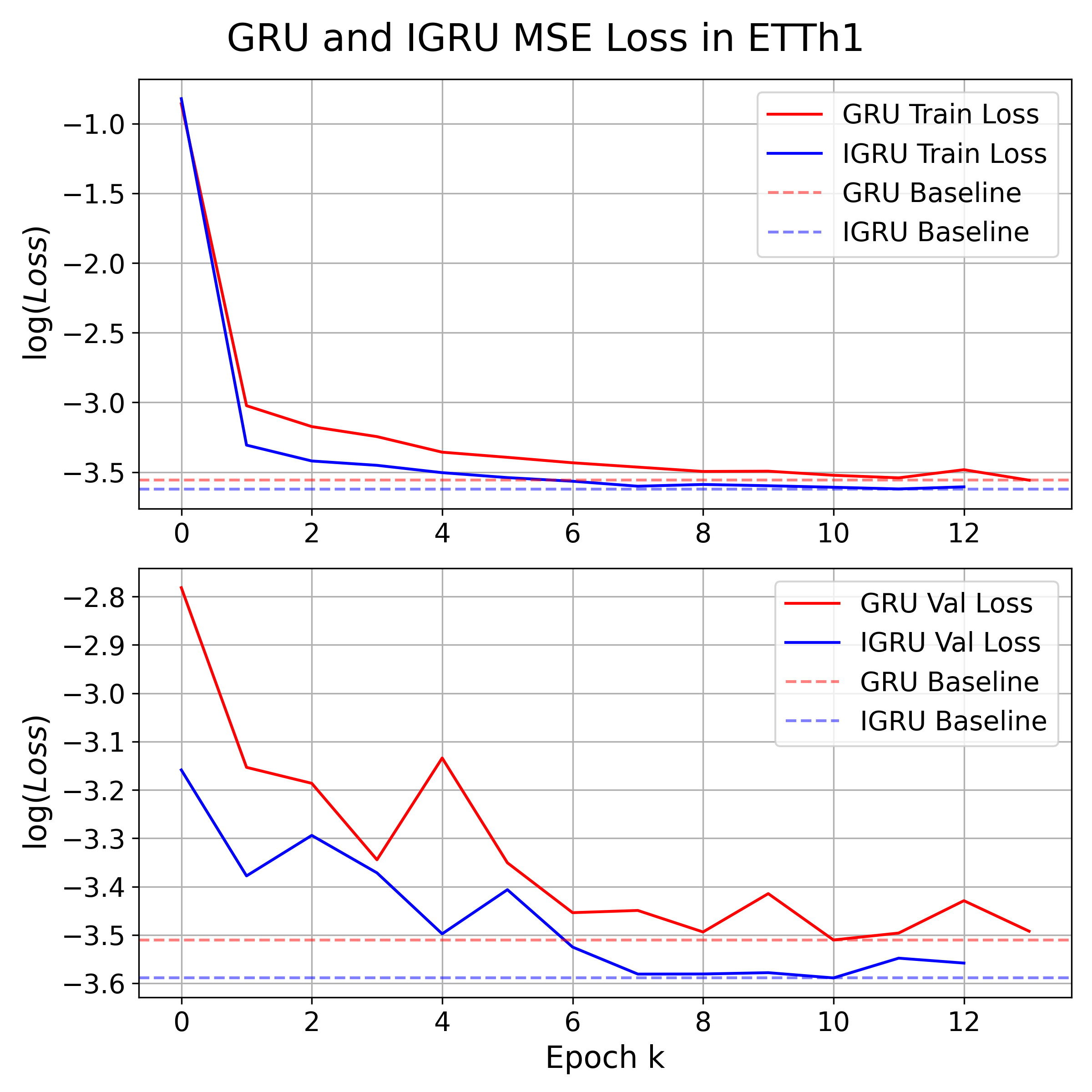}
        \label{fig:loss_GRU_ETTh1}}
    \subfloat[]{
        \includegraphics[width=0.33\textwidth]{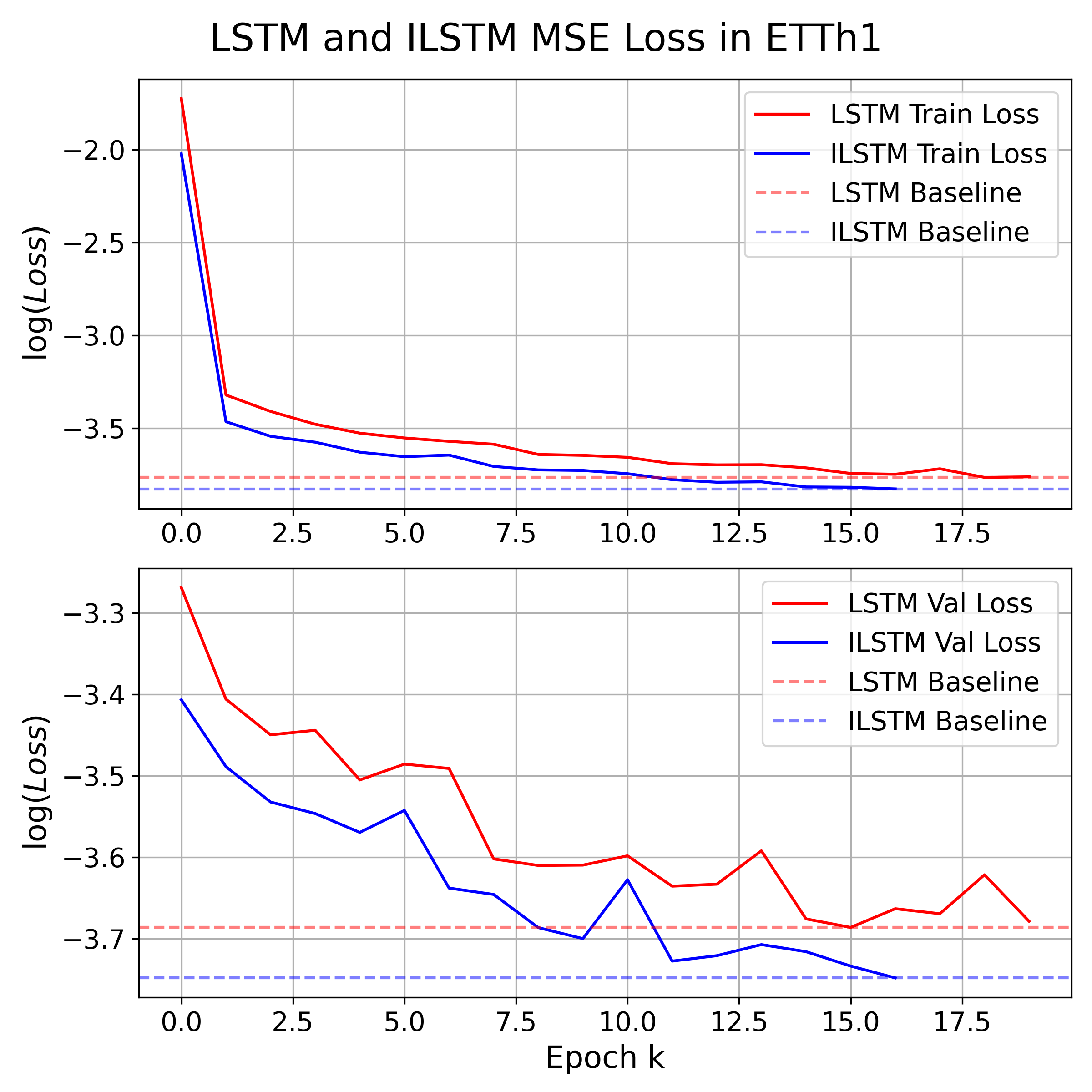}
        \label{fig:loss_LSTM_ETTh1}}
    \caption{Training and validation loss curves on ETTh1. MSE losses are presented on a logarithmic scale. The minimal MSE in each loss curve is marked with a horizontal dashed line. (a) Single-layer RNN and IRNN. (b) GRU and IGRU. (c) LSTM and ILSTM.}
    \label{fig:loss_epoch_ETTh1}
\end{figure*}

In this subsection, we study the performance of IU-BPTT in training IRNNs. Fig. \ref{fig:loss_epoch_ETTh1} displays the training and validation loss curves of IRNNs and their RNN counterparts on ETTh1. Both RNNs trained with BPTT and their IRNN counterparts trained with IU-BPTT show rapid loss declines during the first three epoch, where IRNNs exhibit a faster loss reduction rate on both the training and validation sets. The number of epochs to reach convergence is basically the same for training both RNN and IRNN. This shows the effectiveness of IU-BPTT in training IRNNs.

\begin{figure}[hbtp]
    \centering
    \includegraphics[width=\linewidth]{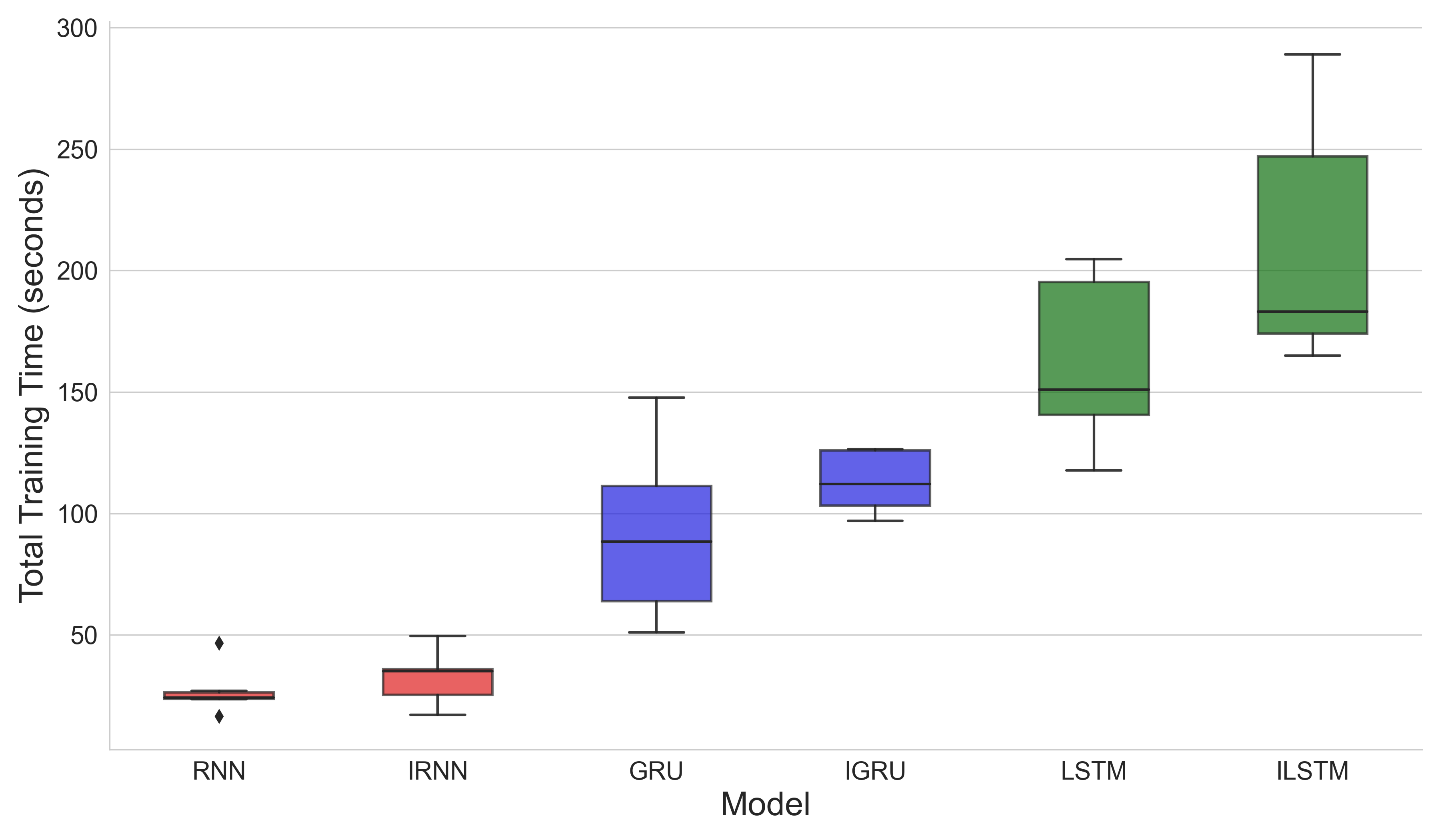}
    \caption{Box plot of total training time for Single-layer IRNN, IGRU, ILSTM, and their RNN counterparts on ETTh1.}
    \label{fig:trainingTime}
\end{figure}

To compare the training cost, we present in Table \ref{tab:epochTime} the average training time per epoch of RNNs and their IRNN counterparts on ETT with 20 random training runs and $N=1$. As a result, the training time of IRNNs includes time spent on both backpropagation and innovation updating. Table \ref{tab:epochTime} shows that training IRNNs with $N=1$ incurs $15\%$ to $20\%$ more time per epoch than training RNNs. Fig. \ref{fig:trainingTime} presents the box plot of total training time on ETTh1. Due to random initialization of network parameters and early stopping strategy (see in Section \ref{sec:implementation}), the total training time varies across different training runs. It can be seen that, on average, training IRNNs require 10\% to 15\% more time compared to training their RNN counterparts without innovations.

\begin{table}[htbp]
\centering
\caption{Average Training Time per Epoch of Six Networks on ETT}
\begin{tabular}{c|cccc}
\toprule
\multirow{2}{*}{\textbf{Network}} 
& \multicolumn{4}{c}{\textbf{Epoch Time (s)}} \\
\cmidrule(lr){2-5}
& ETTh1   & ETTh2   & ETTm1   & ETTm2   \\
\midrule
Single-layer RNN  & 2.1281  & 2.1754  & 8.0280  & 8.3422  \\
Single-layer IRNN & 2.6106  & 2.5812  & 9.5670  & 9.6218  \\
\midrule
GRU          & 3.8472  & 3.5851  & 14.5675 & 14.4850 \\
IGRU         & 4.3717  & 4.4248  & 17.6299 & 17.6468 \\
\midrule
LSTM         & 6.7354  & 7.2820  & 32.7923 & 27.6975 \\
ILSTM        & 7.6865  & 9.0307  & 33.3257 & 30.8128 \\
\bottomrule
\end{tabular}
\label{tab:epochTime}
\end{table}

\begin{figure}[htbp]
    \centering
        \includegraphics[width=\linewidth]{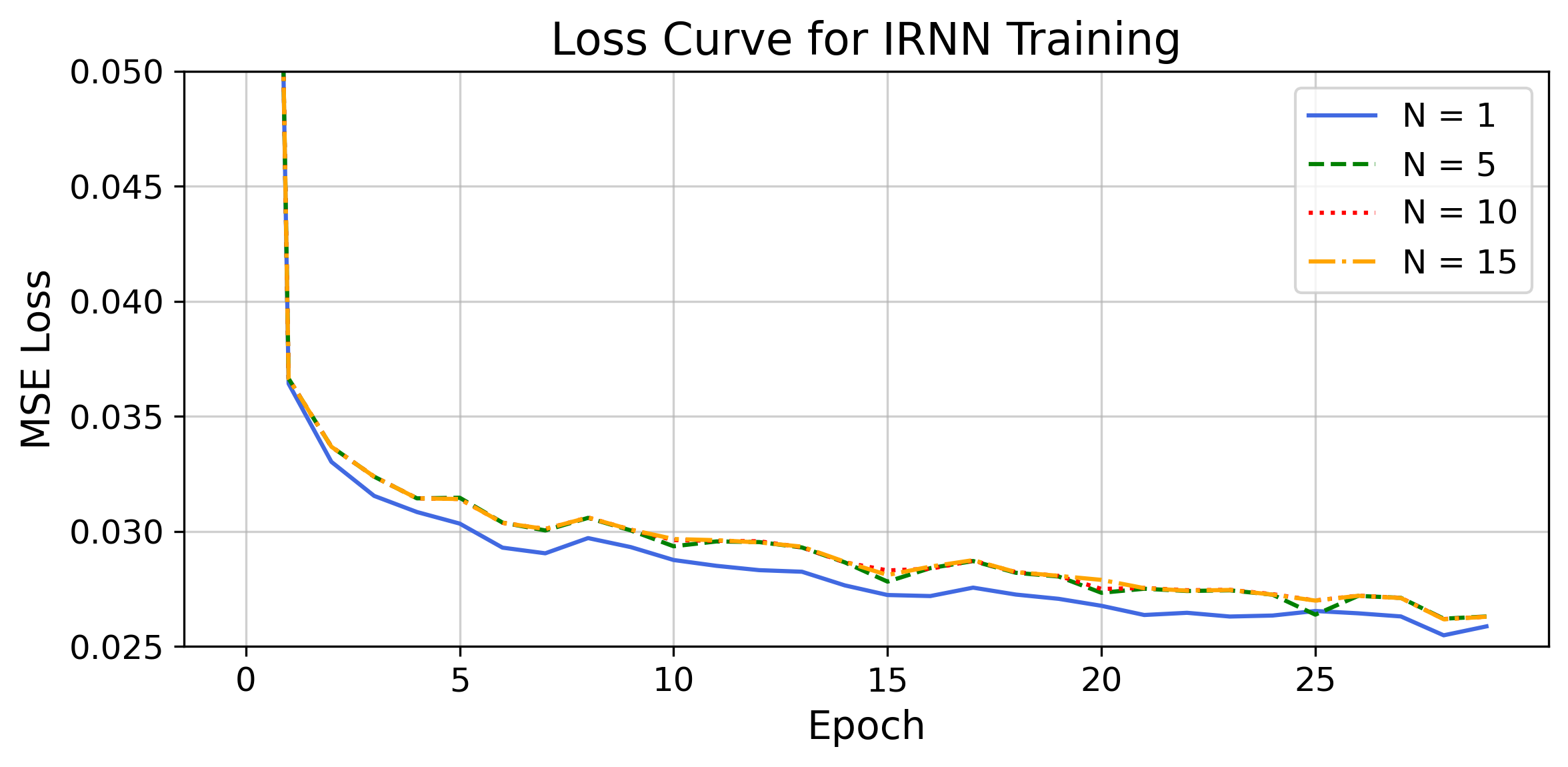}
    \caption{MSE loss curves of Single-layer IRNN with innovation updating intervals $N=1,5,10,15$ on ETTh1 training set.}
    \label{fig:train_loss_N}
\end{figure}

Next, we study the impact of $N$ on training performance. $N$ is set to 1, 5, 10, 15 to train single-layer IRNN, IGRU, and ILSTM. The training of each network with a particular $N$ is repeated 5 times, and the number of epoch in each training is fixed to 30 to evaluate optimization capabilities. The network parameter initialization and dataset loading order are kept the same under different values of $N$. Fig. \ref{fig:train_loss_N} shows the training loss curves of the single-layer IRNN with different $N$. The network training loss of $N=1$ remains the lowest during the whole training process and declines more stably. When a larger $N$ is selected, the efficiency of gradient descent decreases slightly, while still maintaining satisfactory training performance.

Table \ref{tab:epochTime_N} presents the training time per epoch and training MSE of IRNNs on ETTh1 with $N=1,5,10,15$, where experimental results for each $N$ are averaged across 5 training runs. As $N$ increases, the average training MSE of the single-layer IRNN remains nearly unchanged, and that of IGRU and ILSTM increases slightly. On the other hand, the training time per epoch decreases with the increase of $N$, and gradually approaches that of RNN counterparts without innovations. Specifically, when $N=15$, the training time per epoch for the single-layer IRNN, IGRU, and ILSTM increases by only 7.4\%, 2.2\%, and 1.8\%, respectively, compared to that of their RNN counterparts (see in Table \ref{tab:epochTime}). These experimental results reveal that choosing an appropriate large $N$ for IRNN training can save 10\% of training time compared to $N=1$, with only a negligible increase in training MSE. Additionally, the performance of IRNNs trained with larger $N$ is still significantly better than that of their RNN counterparts.

\begin{table}[htbp]
\centering
\begin{threeparttable}
    \caption{Comparison of Training Time and MSE on Training Set for IRNNs with Varying Innovation Updating Intervals $N$ on ETTh1}
    \label{tab:epochTime_N}
    \begin{tabular}{c|ccc}
    \toprule
    Nertwork              & Value of $N$ & Epoch Time (s) & Average MSE \\
    \midrule
    \multirow{4}{*}{\makecell{Single-layer \\ IRNN}} & 1   & 2.6106       & 0.0252  \\
                          & 5   & 2.3856       & 0.0251  \\
                          & 10  & 2.3191       & 0.0253  \\
                          & 15  & 2.2860       & 0.0253  \\
    \midrule                  
    \multirow{4}{*}{IGRU} & 1   & 4.3717       & 0.0281  \\
                          & 5   & 4.0879       & 0.0299  \\
                          & 10  & 4.0135       & 0.0297  \\
                          & 15  & 3.9317       & 0.0297  \\
    \midrule                       
    \multirow{4}{*}{ILSTM} & 1  & 7.6865       & 0.0227  \\
                           & 5  & 7.2168       & 0.0228  \\
                           & 10 & 6.9923       & 0.0229  \\
                           & 15 & 6.8554       & 0.0230  \\
    \bottomrule                       
    \end{tabular}

    \begin{tablenotes}
        \footnotesize
        \item The MSE is averaged across 5 training runs. The epoch time refers to the average across all epochs from 5 training runs.
    \end{tablenotes}
\end{threeparttable}
\end{table}

\subsection{Ablation Studies}
\label{sec:ablation}
For IRNNs with gating mechanisms, it is worth investigating which network modules would benefit more from the introduction of innovations in terms of improving prediction accuracy. Therefore, we conduct two sets of ablation experiments by modifying the network configurations.

\begin{figure}[hbtp]
    \centering
    \subfloat[]{%
        \includegraphics[width=0.25\textwidth]{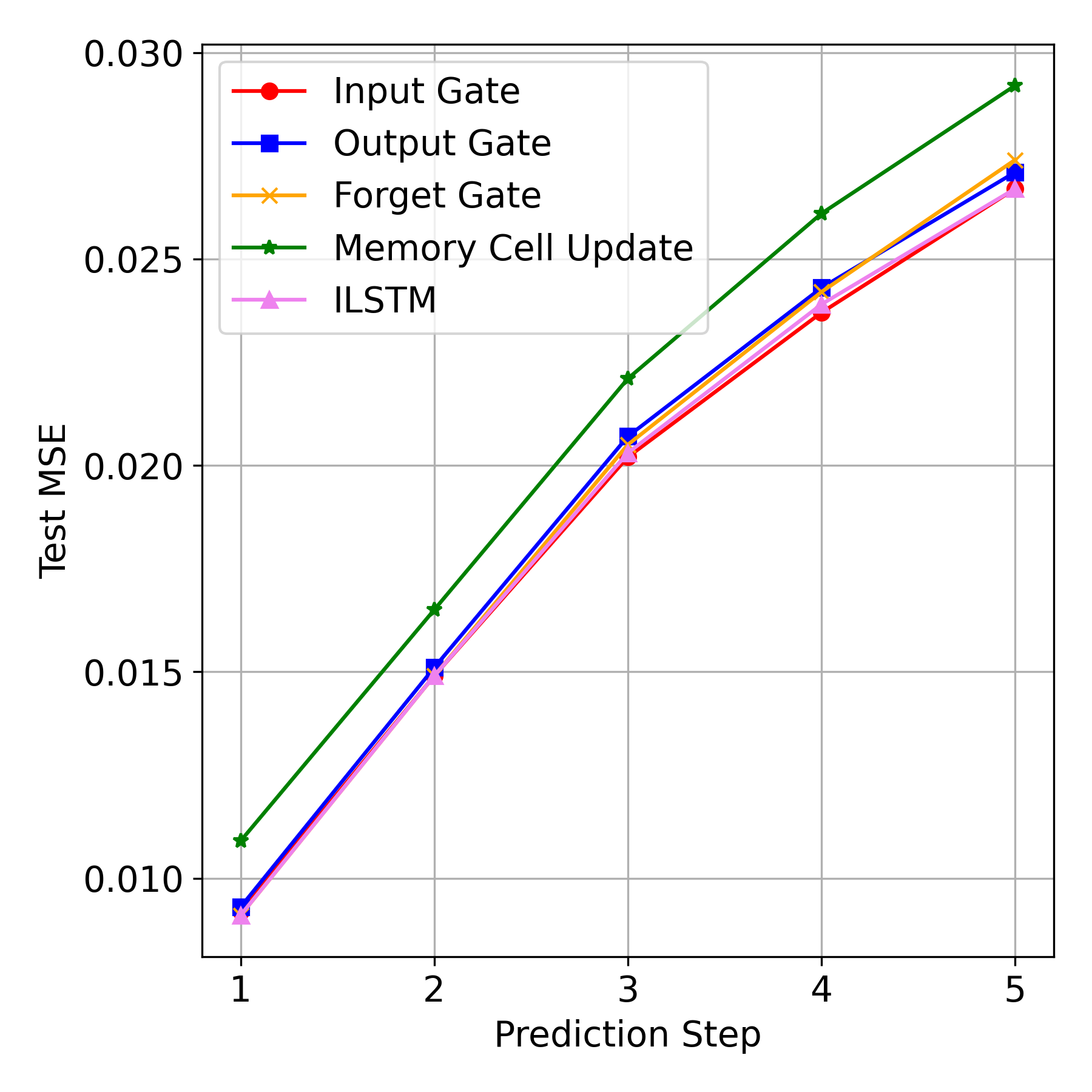}%
        \label{fig:ablation_ILSTM_without}}
    \subfloat[]{%
        \includegraphics[width=0.25\textwidth]{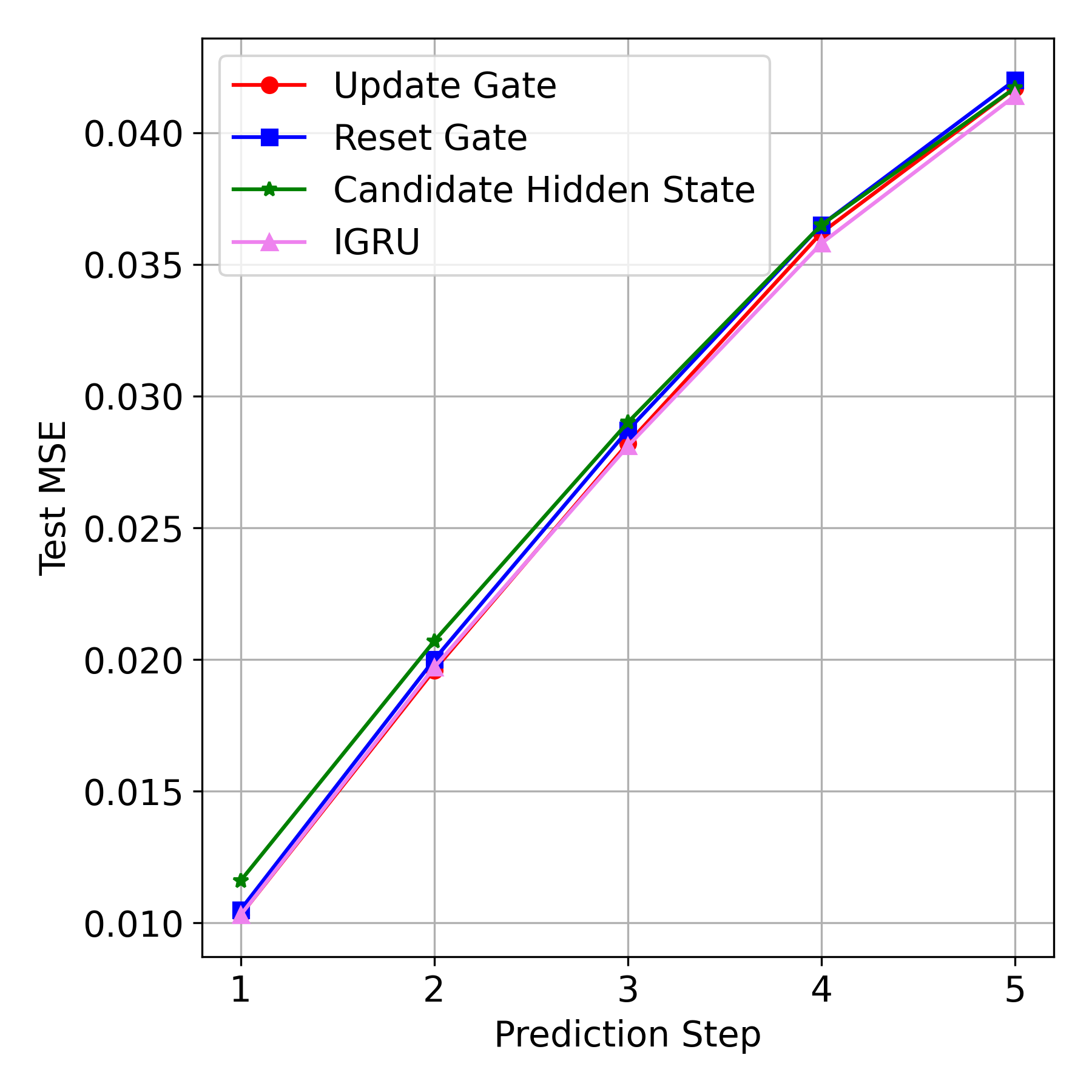}%
        \label{fig:ablation_IGRU_without}}
    \caption{Testing MSE of modified ILSTMs and IGRUs on ETTh1. Each The legend indicates which module excludes innovations. (a) ILSTM and modules without innovation. (b) IGRU and modules without innovation.}
    \label{fig:ablation_without}
\end{figure}

Building upon IGRU and ILSTM as defined in Table \ref{tab:IRNN_equations}, we remove innovations from one specific network module, and keep innovations incorporated into all other modules. These modified networks are trained and compared with IGRU/ILSTM. A higher MSE in these cases indicates that incorporating innovations in the specific module contributes positively to correcting prediction errors. Fig. \ref{fig:ablation_without} presents experiment results on ETTh1 test set. For ILSTM in Fig. \ref{fig:ablation_ILSTM_without}, removing innovations from the memory cell update module $\bm{c}_{t}$ leads to the most significant MSE increase, while removing innovations from the input gate $\bm{g}^{\rm i}_{t}$ makes no difference. When innovations are removed from the forget gate $\bm{g}^{\rm f}_{t}$ or output gate $\bm{g}^{\rm o}_{t}$, the 3-step to 5-step prediction accuracy becomes worse. For IGRU in Fig. \ref{fig:ablation_IGRU_without}, removing innovations from the candidate hidden state $\bm{x}'_{t}$ leads to a noticeable prediction accuracy decline compared to IGRU, especially for 1-step and 2-step prediction. When innovation is removed from update gate $\bm{g}^{\rm u}_{t}$ and reset gate $\bm{g}^{\rm r}_{t}$, the 3-step to 5-step prediction accuracy deteriorates.

\begin{figure}[hbtp]
    \centering
    \subfloat[]{%
        \includegraphics[width=0.25\textwidth]{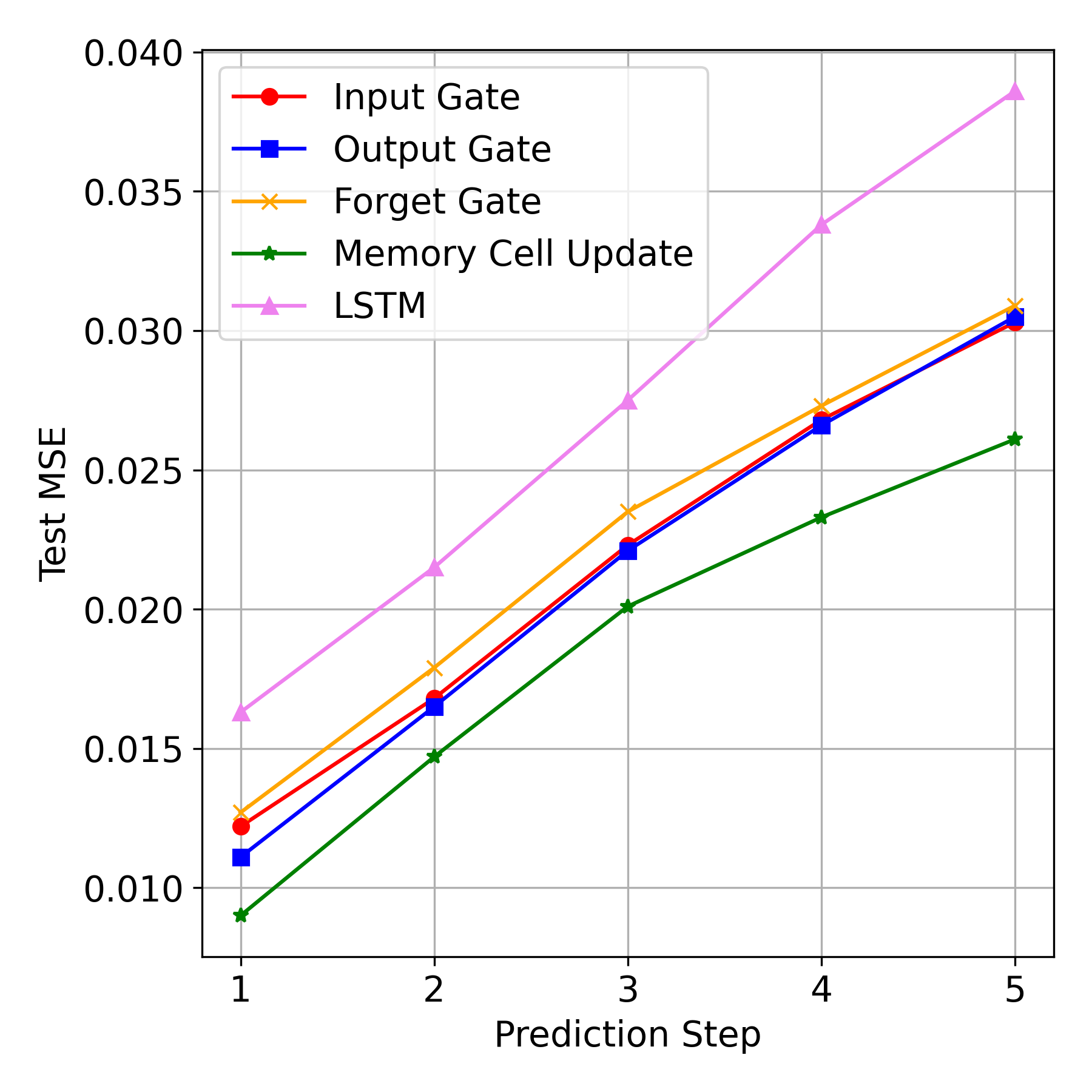}%
        \label{fig:ablation_LSTM_with}}
    \subfloat[]{%
        \includegraphics[width=0.25\textwidth]{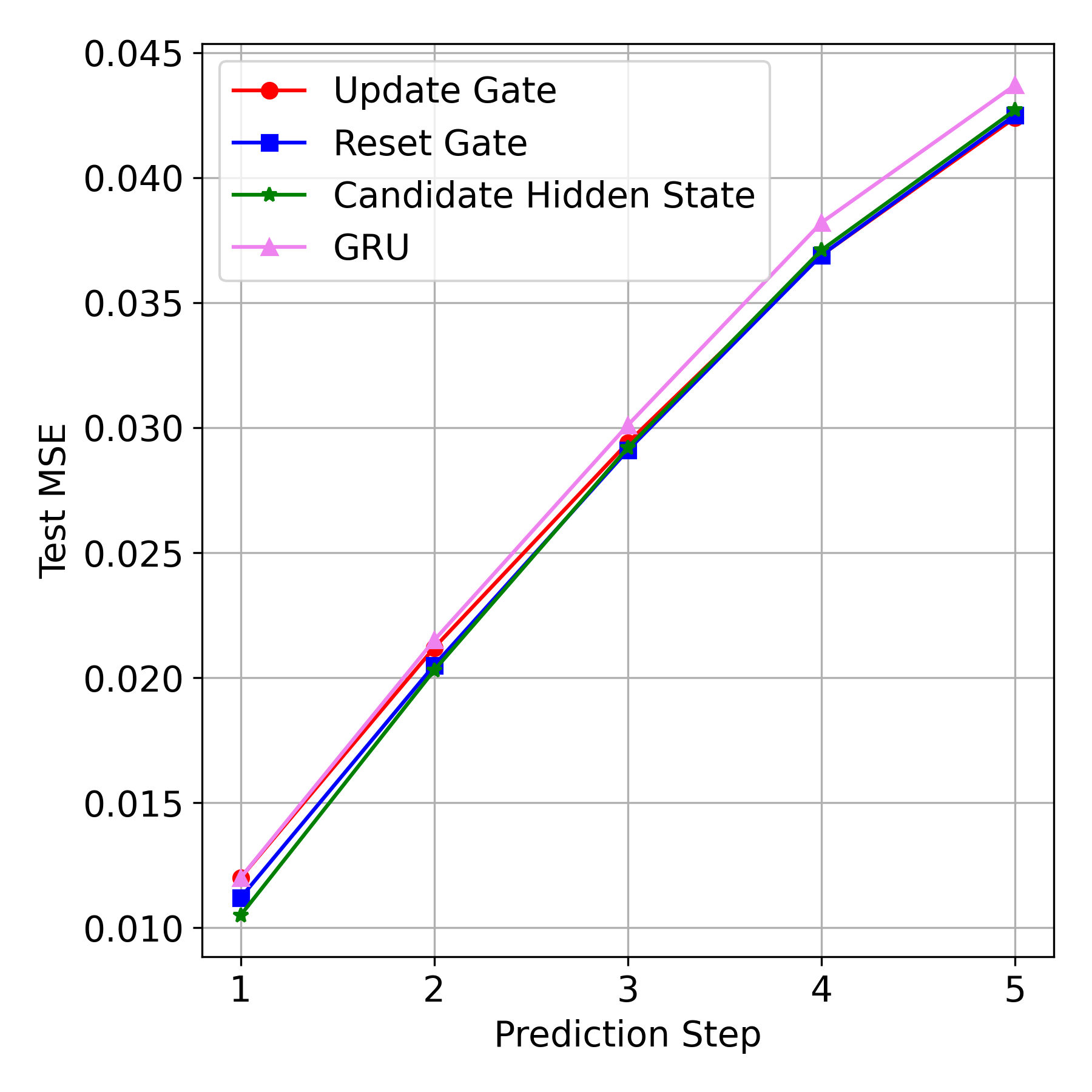}%
        \label{fig:ablation_GRU_with}}
    \caption{Testing MSE of modified LSTMs and GRUs on ETTh1. The legend indicates which module implements innovations. (a) LSTM and modules with innovation. (b) GRU and modules with innovation.}
    \label{fig:ablation_with}
\end{figure}

From another perspective, we incorporate innovations into only one specific network module of GRU/LSTM, and keep other modules unchanged. These modified network are then trained and compared with each other and GRU/LSTM without innovation. A lower MSE in these cases suggests that incorporating innovations into the specific module enhances prediction accuracy. Fig. \ref{fig:ablation_with} shows the experiment results on ETTh1 test set. For LSTM in \ref{fig:ablation_LSTM_with}, incorporating innovations in any module can improve prediction accuracy. Among them, incorporating innovations into $\bm{c}_{t}$ results in the most significant improvement. When innovations are only incorporated into $\bm{g}^{\rm i}_{t}$, $\bm{g}^{\rm o}_{t}$, or $\bm{g}^{\rm f}_{t}$ for LSTM, the three modified networks exhibit comparable prediction accuracy. While their improvement is not as substantial as that achieved by incorporating innovations into $\bm{c}_{t}$, their prediction accuracy is much better than that of LSTM. For GRU, similar results are obtained in Fig. \ref{fig:ablation_GRU_with}. By incorporating innovations into any single module, we always obtain prediction accuracy better than that of GRU. Notably, incorporating innovations in candidate hidden state $\bm{x}'_{t}$ achieves the most significant improvement in prediction accuracy, and incorporating innovations into update gate $\bm{g}^{\rm u}_{t}$ and reset gate $\bm{g}^{\rm r}_{t}$ both improves the 3-step to 5-step prediction accuracy especially.

To summarize, incorporating innovations into $\bm{c}_{t}$ of LSTM and $\bm{x}'_{t}$ of GRU contributes mostly to improving prediction accuracy, and incorporating innovations into gating units is also beneficial.

\section{Conclusion}
\label{sec:conclusion}
In this paper, a new innovation-driven architecture of RNNs was proposed for time-series prediction tasks. By borrowing ideas from KF for LTI systems, we proposed to incorporate innovations into RNN architectures that can be described as nonlinear state-space models. By incorporating innovations into gating mechanism, IGRU and ILSTM were also derived for time-series data modeling and prediction. To adopt the conventional BPTT for training RNNs, a tailored algorithm called IU-BPTT was proposed, which alternates between optimizing network parameters via gradient descent and updating innovations. Extensive experiments conducted on real-world datasets demonstrated that IRNNs can be effectively trained with IU-BPTT and achieve significant accuracy improvements in multi-step prediction compared to their generic counterparts without innovations, at the cost of only a negligible growth in network complexity.

Several directions are worthy of future exploration. An interesting topic is to incorporate innovations into recurrent structure of more advanced RNNs, e.g. the Mamba. Meanwhile, how to integrate innovations with attention mechanisms in Transformers is also a meaningful albeit challenging problem.

\section*{Acknowledgments}
This work was supported by National Natural Science Foundation of China under Grants 62373211 and 62327807.

% reference
\bibliographystyle{IEEEtran}
\bibliography{ref}

\begin{IEEEbiography}[{\includegraphics[width=1in,height=1.25in,clip,keepaspectratio]{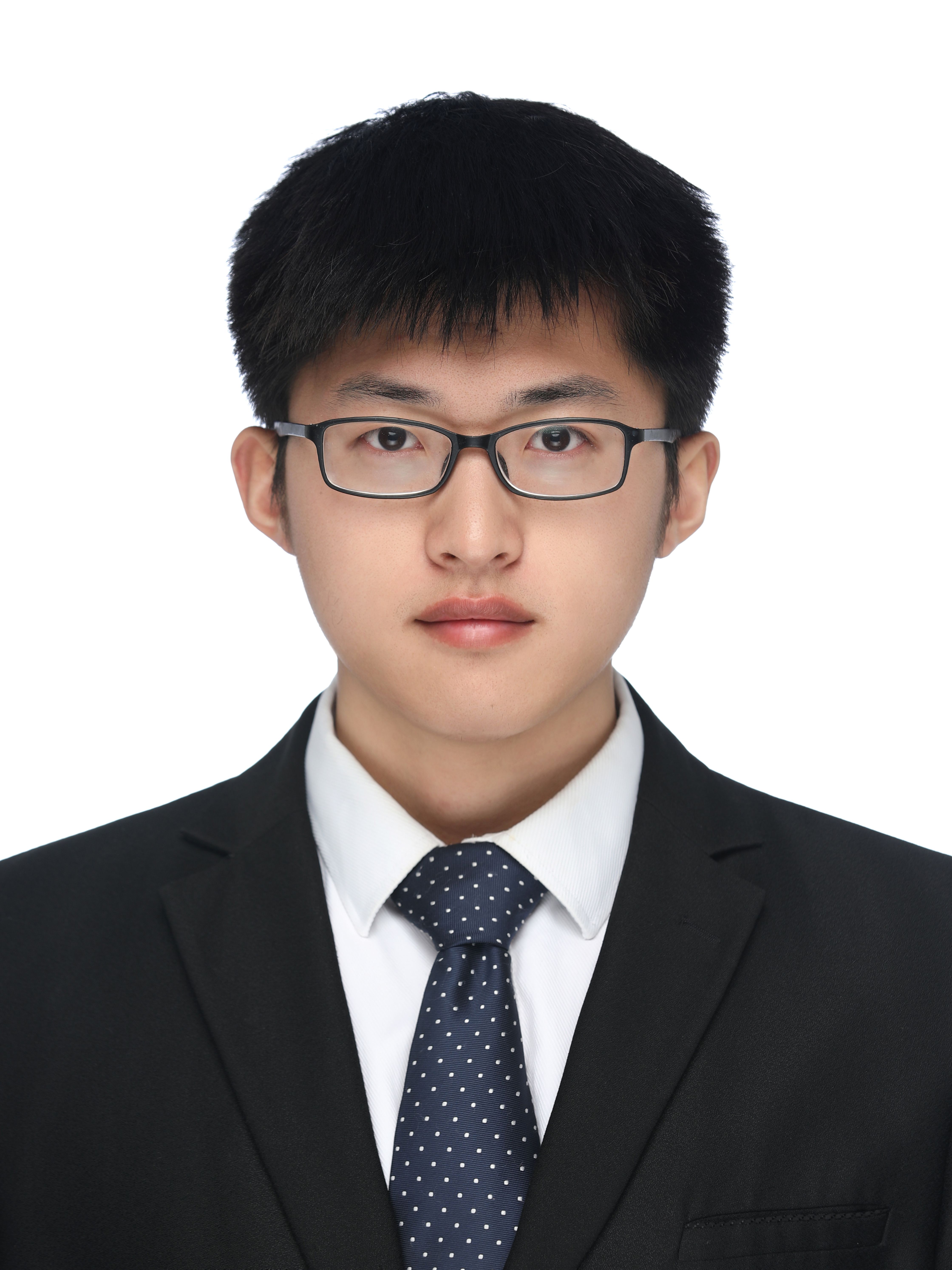}}]{Yifan Zhou} received the B.S. degree in theoretical and applied mechanics from Tsinghua University, Beijing, China, in 2024, where he is currently pursuing the Ph.D. degree in control science and engineering. His current research interests include artificial intelligence techniques and their applications in time series analysis and dynamic system modeling.
\end{IEEEbiography}

\begin{IEEEbiography}[{\includegraphics[width=1in,height=1.25in,clip,keepaspectratio]{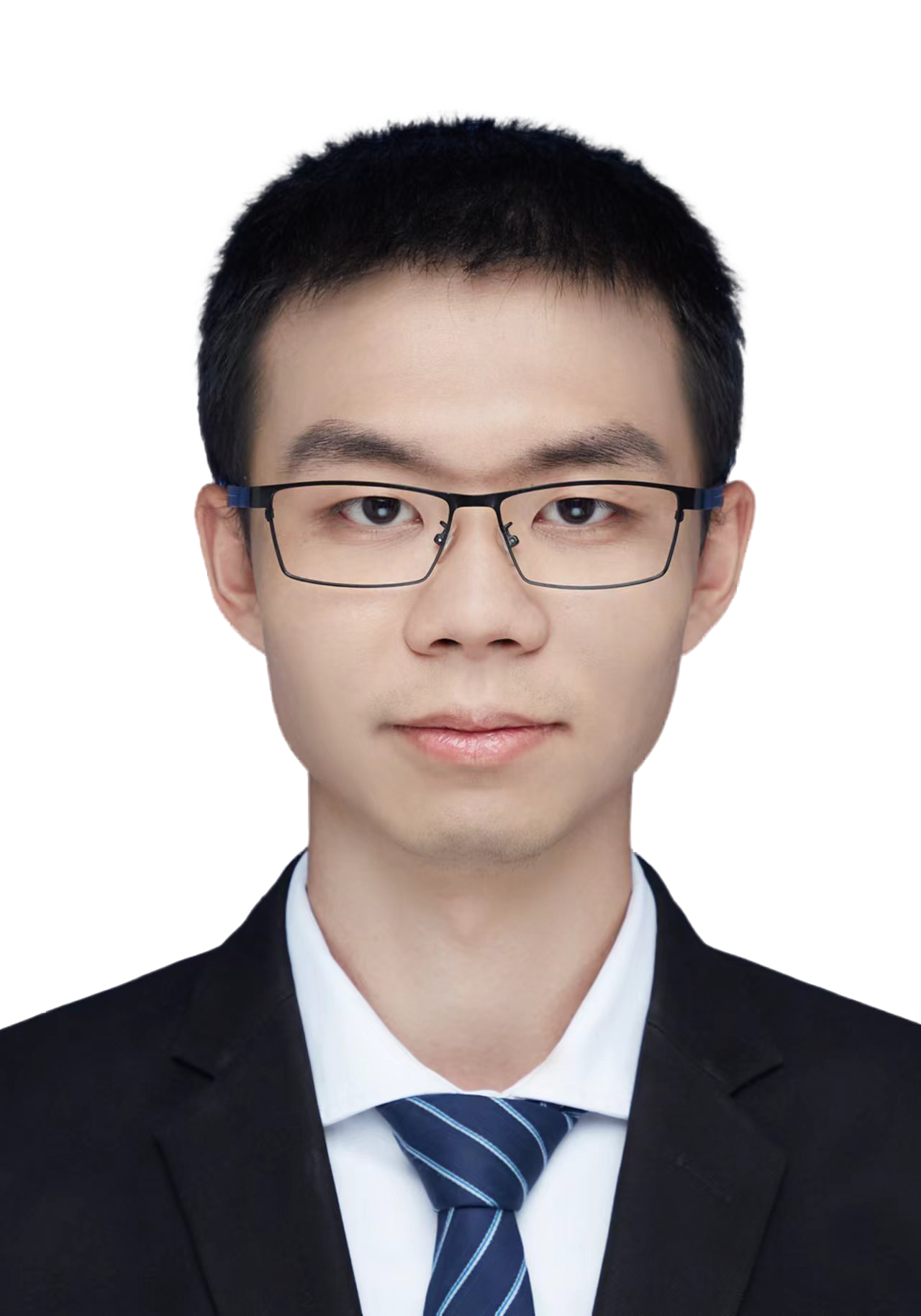}}]{Yibo Wang} received the B.S degree in automation from Tsinghua University, Beijing, China, in 2021, where he is currently pursuing the Ph.D. degree in control science and engineering. His current research interests include data-driven control methods and their applications in industrial processes.

Mr. Wang was a recipient of the Best Student Paper Finalist of the 35th Chinese Process Control Conference.
\end{IEEEbiography}

\begin{IEEEbiography}[{\includegraphics[width=1in,height=1.25in,clip,keepaspectratio]{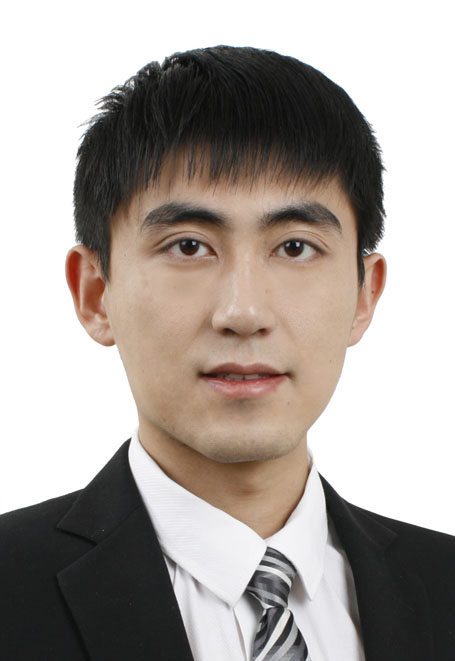}}]{Chao Shang} received the B.Eng. degree in automation and the Ph.D. degree in control science and engineering from Tsinghua University, Beijing, China, in 2011 and 2016, respectively. After working as a Post-Doctoral Fellow at Cornell University, Ithaca, NY, USA, he joined the Department of Automation, Tsinghua University, in 2018, where he is currently an Associate Professor. His research interests range from data-driven modeling, control, and optimization with applications to intelligent manufacturing. 

Prof. Shang was a recipient of the Springer Excellent Doctorate Theses Award, the Emerging Leaders in Control Engineering Practice, the Zhangzhongjun Best Paper Award of Chinese Process Control Conference, and the Best Paper Award of International Conference on Industrial Artificial Intelligence, among others.
\end{IEEEbiography}

\vspace{11pt}

\vfill

\end{document}